\documentclass[10pt, a4paper]{article}

\usepackage{amsmath}
\usepackage{amssymb}
\usepackage{makecell}
\usepackage{booktabs}
\usepackage{listings}
\usepackage{multirow}
\usepackage{multicol}
\usepackage{bm}
\usepackage{hyperref}
\usepackage[capitalize]{cleveref}
\usepackage{twemojis}
\usepackage{xcolor}
\usepackage{xurl}
\usepackage{float}
\usepackage[T1]{fontenc}
\usepackage[utf8]{inputenc}
\usepackage{listings}
\usepackage{footmisc}

\newcommand{\textmod}{\scalebox{1.5}{\twemoji{memo}}}
\newcommand{\audiomod}{\scalebox{1.5}{\twemoji{speaker high volume}}}
\newcommand{\videomod}{\scalebox{1.5}{\twemoji{movie camera}}}
\newcommand{\multimodal}{\scalebox{1}{\twemoji{shuffle tracks button}}}
\newcommand{\imagemod}{\scalebox{1.5}{\twemoji{framed picture}}}
\usepackage{booktabs}
\usepackage{multirow}
\usepackage{xcolor}
\usepackage{colortbl}

\definecolor{sarcblue}{RGB}{180, 210, 240}
\definecolor{nonsarcpink}{RGB}{245, 195, 200}

\newcommand{\lblsarc}{\colorbox{sarcblue}{\texttt{sarc}}}
\newcommand{\lblnonsarc}{\colorbox{nonsarcpink}{\texttt{non-sarc}}}

\usepackage[final]{lrec2026} 

\title{\textsc{MuSaG}: A Multimodal German Sarcasm Dataset with Full-Modal Annotations}

\name{Aaron Scott, Maike Züfle, Jan Niehues} 
\address{Karlsruhe Institute of Technology, Germany \\
         aaron.scott@student.kit.edu,\\
         \{maike.zuefle, jan.niehues\}@kit.edu\\}

\abstract{
Sarcasm is a complex form of figurative language in which the intended meaning contradicts the literal one. Its prevalence in social media and popular culture poses persistent challenges for natural language understanding, sentiment analysis, and content moderation. With the emergence of multimodal large language models, sarcasm detection extends beyond text and requires integrating cues from audio and vision.
We present \textsc{MuSaG}, the first German multimodal sarcasm detection dataset, consisting of 33 minutes of manually selected and human-annotated statements from German television shows. Each instance provides aligned text, audio, and video modalities, annotated separately by humans, enabling evaluation in unimodal and multimodal settings.
We benchmark nine open-source and commercial models, spanning text, audio, vision, and multimodal architectures, and compare their performance to human annotations. Our results show that while humans rely heavily on audio in conversational settings, models perform best on text. This highlights a gap in current multimodal models and motivates the use of \textsc{MuSaG} for developing models better suited to realistic scenarios. 
We release \textsc{MuSaG} publicly to support future research on multimodal sarcasm detection and human–model alignment.
 \\ \newline  }

\begin{document}

\maketitleabstract

\section{Introduction}
\label{sec:introduction}
\footnotetext[1]{\label{fn:data_release}MuSaG is available at \url{https://huggingface.co/datasets/sc0ttypee/MuSaG}}
Sarcasm represents a complex form of figurative language, often conveying meanings that contradict their literal content. The Cambridge Dictionary defines sarcasm as \textit{the use of remarks that clearly mean the opposite of what they say, made in order to hurt someone's feelings or to criticize something in a humorous way}\footnote[2]{\url{https://dictionary.cambridge.org/dictionary/english/sarcasm}}. As such, sarcasm is widespread in user-generated content on social media platforms such as X, Facebook, and Reddit, as well as in popular culture, including sitcoms and movies, where it serves as a key vehicle for humor and mockery \citep{maynard-greenwood-2014-cares}.

Detecting sarcasm is essential for applications such as sentiment analysis \citep{joshi-sarcasm-survey}, hate speech detection \citep{role_of_sarcasm}, and content moderation \citep{DBLP:conf/naacl/LiuXWYYZLGWYMAH25}, since sarcasm can invert the perceived polarity of a statement. With the growing integration of language models into conversational systems, reliable sarcasm detection becomes increasingly important to ensure appropriate and contextually aware responses.
Moreover, with the advent of multimodal large language models \citep{microsoft_phi-4-mini_2025, comanici_gemini_2025, xu_qwen25-omni_2025}, sarcasm detection extends beyond text, requiring understanding across audio and visual modalities as well.

\begin{figure}
    \centering
    \includegraphics[width=1\linewidth]{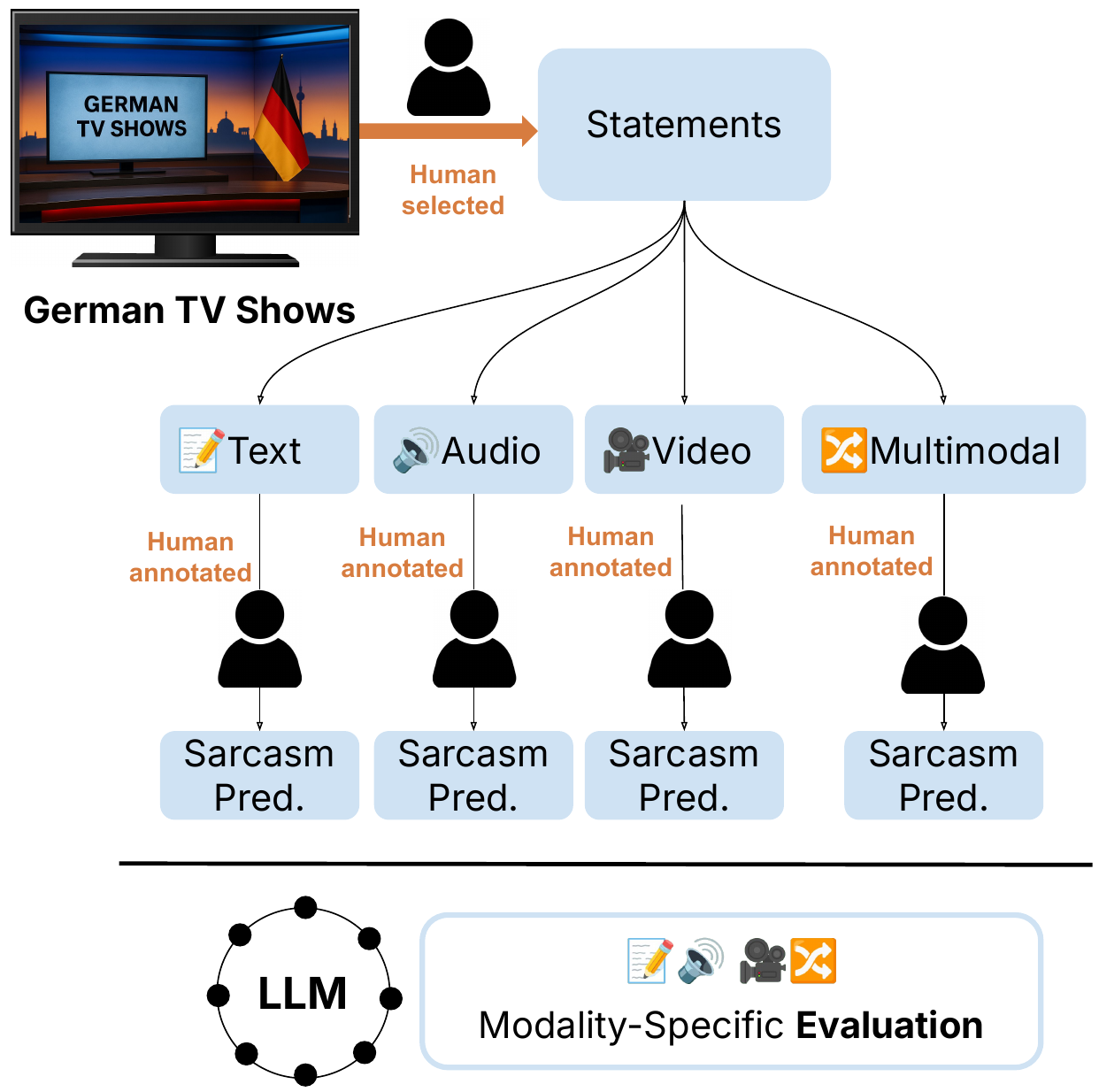}
    \caption{\textsc{MuSaG}, our human annotated German multimodal sarcasm detection dataset.}
    \label{fig:intro_fig}
\end{figure}
\begin{table*}[htbp]
    \resizebox{\linewidth}{!}{%
    \begin{tabular}{p{2.5cm}lccccccccc}
        \toprule
        \textbf{Title} & \textbf{Genre} & \textbf{Lang.} & \textbf{\makecell{\#\\ Srcs}} & \textbf{\makecell{Manual\\ select}} & \textbf{\makecell{Human \\ annot.}} & \textbf{\makecell{Single-Mod.\\ annot.}} & \textbf{\makecell{Agreem. \\ avail.}} & \textbf{\makecell{Text\\ \textmod}} & \textbf{\makecell{Audio\\ \audiomod}}  &   \textbf{\makecell{Vision\\ \videomod \imagemod}} \\
        \midrule
        \citetlanguageresource{cai_multi-modal_2019}*      & Social Media  & en    & 1         & $\times$   & $\times$ & $\times$  & $\times$ & \checkmark & $\times$         & \imagemod      \\
        \citetlanguageresource{schifanella_detecting_2016}  & Social Media  & en    & 3         & $\times$   & \checkmark  & \checkmark & \checkmark    & \checkmark    & $\times$       & \imagemod     \\
        \citetlanguageresource{yue_sarcnet_2024}            & Social Media  & en/zh & 2         & $\times$    & \checkmark  & \checkmark & \checkmark    & \checkmark & $\times$        & \imagemod    \\
        \citetlanguageresource{sangwan_i_2020}              & Social Media  & en    & 1         & $\times$   & (\checkmark)  & \checkmark & \checkmark    & \checkmark & $\times$       & \imagemod      \\
        \citetlanguageresource{alnajjar_que_2021}           & TV-shows      & es    & 2         & \checkmark & \checkmark  & $\times$ & $\times$        & \checkmark & \checkmark    &  \videomod    \\
        \citetlanguageresource{castro_towards_2019}         & TV-shows      & en    & 4         & $\times$   & \checkmark  & $\times$   & \checkmark    & \checkmark & \checkmark   & \videomod    \\
        \citetlanguageresource{zhang_cmma_2023}           & TV-shows      & zh    & 18        & n.s.       & \checkmark  & $\times$   & \checkmark    & \checkmark & \checkmark      & \videomod     \\

        \citetlanguageresource{bedi_multi-modal_2023}       & TV-shows      & hi/en & 1         & n.s.       & \checkmark  & $\times$ & \checkmark      & \checkmark & \checkmark    &  \videomod    \\

        \citetlanguageresource{ray_multimodal_2022}       & TV-shows      & en    & 5         & n.s.       & \checkmark  & $\times$ & \checkmark & \checkmark & \checkmark     & \videomod     \\
        \midrule
        \textsc{MuSaG} (ours)                     & TV-shows      & de    & 4         & \checkmark & \checkmark  & \checkmark & \checkmark  & \checkmark  & \checkmark    &   \videomod     \\
        \bottomrule
        \\
        \multicolumn{8}{l}{* \citetlanguageresource{qin-etal-2023-mmsd2} later manually correct the labels in a derivative of this dataset.}\\
    \end{tabular}%
  }
  \caption{Comparison of available multimodal sarcasm datasets. Papers for which the criterion is fulfilled only for a subset of the data are marked with (\checkmark), and criteria that are not specified are marked as n.s.}
  \label{tab:related_work}
\end{table*}
In text, sarcasm is often indicated by punctuation, hyperbole, or lexical incongruity \citeplanguageresource{tsur_icwsm_2010, davidov_semi-supervised_2010}. In spoken language, prosodic features such as tone, pitch, or emphasis serve as important auditory cues \citeplanguageresource{tepperman_yeah_2006, castro_towards_2019}, while visual expressions, such as smirks or eye-rolls, can also clearly signal sarcastic intent. Accurate sarcasm detection therefore requires integrating cues across modalities and recognizing inconsistencies between them (\citealp{pan_modeling_2020}; \citealplanguageresource{sangwan_i_2020}).

Despite progress in multimodal learning, sarcasm detection remains a challenging task for computational systems \citep{ijcai2024p887}, as successful interpretation depends on subtle contextual, linguistic, and paralinguistic information. A key limitation is that most existing multimodal sarcasm datasets are in English \citep{ijcai2024p887}, although sarcasm is a pervasive, multilingual phenomenon. Moreover, existing resources rarely support modality-specific evaluation.

To address this gap, we introduce \textsc{MuSaG}, a German multimodal sarcasm detection dataset comprising 33 minutes of human-annotated statements from German television shows. Each statement has been manually selected rather than relying on automatically tagged data \citeplanguageresource{schifanella_detecting_2016, cai_multi-modal_2019, castro_towards_2019, sangwan_i_2020}. Each instance includes aligned text, audio, and video modalities, all separately annotated by humans, enabling evaluation in multimodal and unimodal settings (text-only, audio-only, vision-only, and their combinations). This is visualized in \cref{fig:intro_fig}.

We benchmark nine open-source and commercial models, three text-based, one audio-based, two vision-based, and three multimodal systems, to examine their ability to detect sarcasm and compare their predictions with human annotations.  We find that audio provides the strongest unimodal cues for humans, followed by text and then video. In contrast, models perform best on text, indicating that current multimodal systems still struggle to effectively integrate non-textual information. This highlights a gap between text-based model performance and real conversational sarcasm. Furthermore, we analyze the effect of adding broader conversational context and observe that it does not improve performance but instead degrades the models' effectiveness, thereby limiting its usefulness in real-world scenarios.

Our main contributions are as follows:
\begin{enumerate}
    \item We release the first open, human-annotated German multimodal sarcasm dataset with modality-specific annotations.\textsuperscript{\ref{fn:data_release}}
    \item We evaluate nine state-of-the-art unimodal and multimodal models, both commercial and open-source, across modality configurations.
    \item We show that in contrast to humans, current multimodal models fail to leverage audio and visual cues, instead relying primarily on text.
\end{enumerate}

\section{Related Work}\label{sec:rel_work}
We briefly look into unimodal datasets, before then discuss multimodal datasets.

\paragraph{Unimodal Sarcasm Detection}
The importance of sarcasm detection was recognized early, leading to the development of several text-based benchmarks \citeplanguageresource[among others]{tsur_icwsm_2010, davidov_semi-supervised_2010, gonzalez-ibanez_identifying_2011, wallace_humans_2014}. These datasets were primarily constructed from social media platforms such as Twitter or from product reviews, focusing on lexical and syntactic cues for sarcasm.  

Audio-based sarcasm detection has also been explored, with datasets leveraging prosodic and intonational features directly from speech \citeplanguageresource{tepperman_yeah_2006} or indirectly through transcribed TV dialogues \citeplanguageresource{joshi_harnessing_2016}.

\paragraph{Multimodal Sarcasm Detection}
\cref{tab:related_work} provides an overview of publicly available multimodal sarcasm detection datasets and compares them with our proposed \textsc{MuSaG} corpus.
Earlier resources primarily focus on social media content, combining text with accompanying images or metadata \citeplanguageresource{schifanella_detecting_2016, cai_multi-modal_2019, sangwan_i_2020, yue_sarcnet_2024}.
More recent datasets based on television material \citeplanguageresource{castro_towards_2019, ray_multimodal_2022, bedi_multi-modal_2023, zhang_cmma_2023} introduce aligned audio–visual components, yet often lack fine-grained modality separation or manual selection of source clips.

Most existing datasets are in English, with only three multilingual exceptions: English–Chinese \citeplanguageresource{yue_sarcnet_2024}, Hindi–English \citeplanguageresource{bedi_multi-modal_2023}, and Spanish \citeplanguageresource{alnajjar_que_2021}.
Among these, only \citetlanguageresource{alnajjar_que_2021} does not rely on automatically collected data.
While several datasets include human annotations, only a subset, limited to text-image datasets, provides modality-specific labels \citeplanguageresource{schifanella_detecting_2016, sangwan_i_2020, yue_sarcnet_2024}.

To date, no dataset provides full multimodal coverage with modality-specific annotations, an essential requirement for analyzing how multimodal conversational models interpret sarcasm.
In contrast, \textsc{MuSaG} offers manually curated, human-annotated German data with independent annotations across all modalities.

\section{Dataset}
\label{sec:dataset}
We present \textsc{MuSaG}, a manually curated German multimodal sarcasm dataset enabling analysis across text, audio, and video modalities.
This section details the dataset’s collection, processing and human annotation, as well as dataset statistics. The dataset will be released on HuggingFace upon paper acceptance.

\subsection{Data Collection}
\label{subsec:data_collection}
We selected four German TV shows known for their explicitly sarcastic style and officially described as such by their producers: \textit{Reschke Fernsehn}\footnote{\small{\url{https://www.ardmediathek.de/sendung/reschke-fernsehen/Y3JpZDovL2Rhc2Vyc3RlLm5kci5kZS80ODY3}}}, \textit{heute show}\footnote{\url{https://www.zdf.de/shows/heute-show-104}}, \textit{Die Carolin Kebekus Show}\footnote{\url{https://www.ardmediathek.de/sendung/die-carolin-kebekus-show/Y3JpZDovL2Rhc2Vyc3RlLmRlL2RpZS1jYXJvbGluLWtlYmVrdXMtc2hvdw}}, and \textit{extra 3}\footnote{\url{https://www.ndr.de/fernsehen/sendungen/extra_3}}. All shows are part of official German public broadcasting productions and provide publicly accessible video recordings. We include videos released after April 2024.

From these sources, we manually collected a balanced set of candidate statements to ensure coverage across speaker gender and potential sarcastic content.
In an initial selection phase, we identified short segments that appeared likely sarcastic or clearly non-sarcastic, capturing a range of expressions from overt to subtle. Final sarcasm labels were determined through a subsequent human annotation process as detailed in \cref{subsec:human_annotation}.
\begin{table*}[ht]
    \centering
    \resizebox{\linewidth}{!}{%
    \begin{tabular}{llcccccc}
    \toprule
        & & \textbf{\# State-} & \textbf{\# Female} & \textbf{\# Male} & \textbf{Total Video} &  \textbf{Avg. Video} & \textbf{Avg. Words}\\
        & & \textbf{ments} & \textbf{Speakers} & \textbf{Speakers} & \textbf{min} & \textbf{sec} & \textbf{in Transcript} \\

    \midrule
    \multirow{3}{*}{\textsc{MuSaG} }& Sarcastic   & 120 & 53 & 57 & 19.12 & 9.56 $\pm$ 4.68 & 24.15 $\pm$ 11.32 \\
     & Non-Sarcastic & 94 & 53 & 50 & 13.55 & 8.65 $\pm$ 2.89 & 21.18 $\pm$ 7.78 \\
    \cmidrule(lr){2-8}
    & Total & 214 & 107 & 107 & 32.68 & 9.16 $\pm$ 4.02 & 22.85 $\pm$ 10.03 \\
    \midrule
    \multirow{3}{*}{\textsc{MuSaG-FullAgree}} & Sarcastic   & 96 & 47 & 49 & 14.93 & 9.33 $\pm$ 4.80 & 23.65 $\pm$ 11.48 \\
    & Non-Sarcastic & 59 & 38 & 21 & 8.75 & 8.89 $\pm$ 2.71 & 22.25 $\pm$ 7.87 \\
    \cmidrule(lr){2-8}
    & Total & 155 & 85 & 70 & 23.67 & 9.16 $\pm$ 4.13 & 23.12 $\pm$ 10.28 \\
    \bottomrule
    \end{tabular}%
    }
    \caption{Dataset statistics for our German multimodal dataset \textsc{MuSaG} and the variant \textsc{MuSaG-FullAgree}, that has full annotator agreement.}
    \label{tab:dataset_statistics}
\end{table*}
\subsection{Data Processing}
\label{subsec:data_processing}
To prepare the collected segments for multimodal analysis, we downloaded the videos and split the audio and video streams into separate files. Audio files were sampled at 44.1 kHz with a bitrate of 320 kbps, while video files were downsampled to 426×240 pixels at 15 frames per second. The authors manually verified that this resolution and frame rate preserved key visual cues, including facial expressions and gestures.

To provide corresponding textual data, the audio was automatically transcribed using OpenAI Whisper 3 \citep{radford2022whisper}, and the transcripts were subsequently post-edited by a human annotator with native German proficiency and expertise in multimodal analysis.

\subsection{Human Annotation}
\label{subsec:human_annotation}
\paragraph{Multimodal Annotation.}
To label the dataset, we designed a human annotation process involving 12 participants with strong German language proficiency—11 native speakers and one highly proficient non-native speaker. Each annotator was assigned a subset of the dataset and asked to assign a label (‘sarc’ or ‘non-sarc’) based on the audiovisual representation of each statement, reflecting real conversational conditions. Annotators could also leave comments in case of technical issues. Each statement was annotated by three annotators, and the final label was determined using a majority vote.
The inter-annotator agreement, measured using Fleiss’ Kappa, is 0.623, indicating substantial agreement \citep{landis_measurement_1977}.


Detailed instructions provided to the annotators are included in \cref{fig:instructions_german} in \cref{app:human_annotation_instructions} (in German), with an English translation in \cref{fig:instructions_english}.

\paragraph{Single Modality Annotation.}
In addition to multimodal labels, we obtained human annotations for isolated modalities as part of the initial annotation task. To avoid bias, each annotator classified a statement in only one modality. Annotators used the same processed data that were later provided to LLMs for classification, enabling direct comparison between human and model performance.

To ensure comparability, annotators were instructed to read, watch, or listen to each statement only once. The inter-annotator agreement for single-modality annotations is slightly lower at 0.594. 
Detailed instructions to the annotators are provided in \cref{app:human_annotation_instructions}.

\subsection{Dataset Statistics}
\label{subsec:datset_statistics}
Our final dataset contains 214 statements, of which 120 are sarcastic and 94 are non-sarcastic. Speaker gender is perfectly balanced, with 107 statements each from female and male speakers. On average, statements contain 22.85 words and are spoken over 9.16 seconds.

The dataset includes three modalities: audio, video, and transcript, with transcripts manually reviewed and corrected. In addition, we release the individual annotations from all human annotators, enabling detailed analysis of agreement and variability. Among the statements, 155 of 214 have full agreement across annotators; we refer to this subset as \textsc{MuSaG-FullAgree}, representing entries with unanimous labels for the audio-video modality.

Comprehensive statistics for both the full dataset and \textsc{MuSaG-FullAgree} can be found in \cref{tab:dataset_statistics}.

\section{Analysis}
We now benchmark a range of state-of-the-art open-source and commercial models, both unimodal and multimodal, on the \textsc{MuSaG} dataset. Our analysis examines how well models detect sarcasm across different input modalities and how closely their predictions align with human judgments.

\subsection{Experiment Setting}
\label{subsec:experimental_settings}
We assess model performance using precision, recall, and F1-score across multiple modality configurations to measure their ability to detect sarcasm from text, audio, and visual information.

For unimodal evaluation, text, audio, and video models are tested on their respective input types (text-only, audio-only, and video-only). To explore potential cross-modal benefits, we additionally evaluate unimodal models with the inclusion of textual input, i.e., audio-text and video-text configurations.
Multimodal models are evaluated on single modalities as well as on all available combinations, including text–audio, text–video, and the most realistic, human-like condition: audio–video.

Each model is instructed to classify a statement as either sarcastic or non-sarcastic. We use two prompting strategies, both in English:
\begin{enumerate}
    \item Generic prompt: \textit{“Decide based on the input whether the given utterance is sarcastic or not sarcastic. Answer only with 'sarc' or 'non-sarc'.”}
    \item Modality-specific prompt: The basic prompt with an addition to describe the input format (speech, video, or transcript). For example for the transcript, the addition is \textit{“You will be given one short sentence at a time, containing the transcript of a spoken statement.”}
\end{enumerate}
\begin{table*}[ht]
    \centering
    \begin{tabular}{llccc}
    \toprule
    \textbf{Modality} & \textbf{Model} & \textbf{Precision} & \textbf{Recall} & \textbf{F1} \\
    \midrule
    N/A & random baseline & 52.15 & 52.15 & 52.15\\
    \midrule
    \multirow{6}{*}{text \textmod{}} 
     & \textmod{} Qwen2-7B-Instruct & 76.69 & 64.82 & 62.83 \\ 
     & \textmod{} Qwen2.5-7B-Instruct & 75.11 & 71.18 & 71.33 \\
     & \textmod{} Qwen3-8B & 83.32 & 83.24 & \textbf{83.28} \\
     & \multimodal{} Phi-4-multimodal-instruct & 65.79 & 65.62 & 65.68 \\
     & \multimodal{} Qwen2.5-Omni-7B & 79.01 & 64.59 & 62.14\\
     & \multimodal{} Gemini-2.5-flash & \textbf{85.34} & \textbf{83.75} & 81.71 \\
    
    \midrule 
     \multirow{4}{*}{audio \audiomod{}}     
     & \audiomod{} Qwen2-Audio-7B- Instruct & 58.38 & 57.65 & 55.18 \\ 
     & \multimodal{} Phi-4-multimodal-instruct & 55.17 & 53.76 & 48.28 \\
     & \multimodal{} Qwen2.5-Omni-7B & \textbf{80.63} & 67.78 & 66.45 \\
     & \multimodal{} Gemini-2.5-flash & 79.01 & \textbf{71.67} & \textbf{66.95} \\ 
     \midrule
     \multirow{5}{*}{video \videomod{}} & \videomod{} Qwen2-VL-7B-Instruct & 56.66 & 56.76 & 56.43 \\
     & \videomod{} Qwen2.5-VL-7B-Instruct & 60.23 & 52.87 & 39.65 \\
     & \multimodal{} Phi-4-multimodal-instruct & 21.96 & 50.00 & 30.52 \\
     & \multimodal{} Qwen2.5-Omni-7B & \textbf{61.23} & 59.29 & 55.48 \\
    & \multimodal{} Gemini-2.5-flash & 60.59 & \textbf{60.74} & \textbf{60.53} \\

     \midrule
     \multirow{4}{*}{text-audio \textmod{}\audiomod{}}     &  \audiomod{} Qwen2-Audio-7B- Instruct & 61.50 & 60.50 & 58.01 \\ 
     & \multimodal{} Phi-4-multimodal-instruct & 46.89 & 49.33 & 34.41 \\
      & \multimodal{} Qwen2.5-Omni-7B & 79.29 & 65.12 & 62.88 \\
     & \multimodal{} Gemini-2.5-flash & \underline{\textbf{87.47}} & \underline{\textbf{87.87}} & \underline{\textbf{86.91}} \\
     \midrule 

     \multirow{5}{*}{text-video \textmod{}\videomod{}}    & \videomod{} Qwen2-VL-7B-Instruct & 66.00 & 61.30 & 59.95 \\
     & \videomod{} Qwen2.5-VL-7B-Instruct & 74.92 & 75.15 & 74.99 \\
     & \multimodal{} Phi-4-multimodal-instruct & 63.79 & 52.75 & 37.33 \\
     & \multimodal{} Qwen2.5-Omni-7B & 79.84 & 66.19 & 64.33 \\
    & \multimodal{} Gemini-2.5-flash & \textbf{84.09} & \textbf{84.49} & \textbf{83.63} \\
     \midrule 
     \multirow{3}{*}{audio-video \audiomod{}\videomod{}} & \multimodal{} Phi-4-multimodal-instruct & 62.55 & 52.27 & 37.03 \\
     & \multimodal{} Qwen2.5-Omni-7B & \textbf{76.45} & 66.00 & 64.55 \\
    & \multimodal{} Gemini-2.5-flash & 74.87 & \textbf{74.92} & \textbf{74.89} \\
     \midrule
     \multirow{3}{*}{text-audio-video \textmod{} \audiomod{}\videomod{}} & \multimodal{} Phi-4-multimodal-instruct & 62.34 & 59.26 & 54.17 \\
     & \multimodal{} Qwen2.5-Omni-7B & 81.37 & 72.80 & 72.79 \\
    & \multimodal{} Gemini-2.5-flash & \textbf{83.42} & \textbf{83.34} & \textbf{83.38} \\
     \bottomrule
    \end{tabular}
    \caption{Results on our newly proposed \textsc{MuSaG} dataset for different modalities. We report the macro average over the sacrcastic and non-sarcastic class. For each modality, we report modality-specific models (\textmod{}, \audiomod{}, \videomod{}) and multimodal models (\multimodal{}).}
    \label{tab:results}
\end{table*}
For each model, we report results corresponding to the prompting strategy that yielded the best performance. 
Full prompt templates and model-specific settings are provided in \cref{app:result_config}.

\subsection{Extended Context}
\label{subsec:extended_context_setup}
To investigate the effect of additional context on classification performance, we include up to 15 seconds of preceding content for each statement. This duration was chosen based on our dataset statistics (\cref{tab:dataset_statistics}), which show that individual statements are on average around 10 seconds long, meaning 15 seconds reliably captures at least one additional preceding utterance. Example utterances with extended context are shown in Table~\ref{tab:context_effects} in Appendix~\ref{app:examples}. 
Prompts are constructed to provide this extended context to the model, with the target statement explicitly indicated in the prompt using its transcript.
Full prompts for the extended context condition are provided in \cref{apdx:extended_context}.

\begin{table*}[tp]
\centering
\resizebox{\textwidth}{!}{%
\renewcommand{\arraystretch}{1.6}
\begin{tabular}{l l p{6.5cm} c p{4.5cm}}
\toprule
\textbf{Modality} & \textbf{Model} & \textbf{Transcript} & \textbf{True Label} & \textbf{Disagreement} \\
\midrule

\multirow[c]{2}{*}{Audio} & \multirow[c]{2}{*}{Gemini-2.5-flash} &
\textit{„Ist jetzt vielleicht auf Anhieb nicht so schnell zu erkennen, es sind 83 Prozent Wirtschaft in einem Ausschuss. Und wenn ich das richtig überblicke, sind das deutlich mehr als 50 Prozent."}
& \lblnonsarc & H: \lblsarc \quad M: \lblnonsarc \\[6pt]

& &
\textit{„Ewige Chemikalien, man findet sie überall, im Regen, in menschlichem Blut, selbst in der Arktis wurde PFAS jetzt schon nachgewiesen, in der Arktis."}
& \lblnonsarc & H: \lblnonsarc \quad M: \lblsarc \\

\midrule

\multirow[c]{2}{*}{Video} & \multirow[c]{2}{*}{Gemini-2.5-flash} &
\textit{„In der Regel. Was heißt denn das? Immer? Manchmal? Nur wenn die Sonne in Konjunktion mit Jupiter steht?"}
& \lblsarc & H: \lblsarc \quad M: \lblnonsarc \\[6pt]

& &
\textit{„Also, gerade kleine und mittelständische Unternehmen leiden wohl besonders unter der Bürokratie."}
& \lblnonsarc & H: \lblnonsarc \quad M: \lblsarc \\

\midrule

\multirow[c]{2}{*}{Text} & \multirow[c]{2}{*}{Qwen3-8B} &
\textit{„Söder hat aktuell ein bisschen Zoff mit dem CDU-Mann Daniel Günther. Günther sagte, Söder solle nicht die ganze Zeit von Schwarz-Grün reden. Söder hat seriös und ohne persönlich zu werden geantwortet."}
& \lblsarc & H: \lblsarc \quad M: \lblnonsarc \\[6pt]

& &
\textit{„Why? Deutschlands Behörden, ist kein Witz, nutzen aktuell um die 10.000 verschiedene Softwarelösungen."}
& \lblnonsarc & H: \lblnonsarc \quad M: \lblsarc \\

\bottomrule
\end{tabular}
}%
\caption{Examples of human--model disagreements per modality. For each modality, we show two cases: one where humans rated an utterance as sarcastic while the model did not (H:\,\lblsarc, M:\,\lblnonsarc), and one where the model predicted sarcasm while humans did not (H:\,\lblnonsarc, M:\,\lblsarc). H\,=\,human annotators, M\,=\,model prediction.}
\label{tab:disagreements_modality}
\end{table*}

\subsubsection{Models}
We benchmark nine different models on \textsc{MuSaG}, comprising eight open-source and one commercial model, Gemini \citep{comanici_gemini_2025}. The selection includes three text-based LLMs, one audio model, two vision models, and three fully multimodal LLMs, as detailed below:

\begin{itemize}
    \item \textbf{\textmod{} Text-based LLMs:} Qwen3-8B \citep{yang_qwen3_2025}, Qwen2.5-7B-Instruct \citep{qwen_qwen25_2025}, Qwen2-7B-Instruct \citep{yang_qwen2_2024}.
    \item \textbf{\audiomod{}\videomod{} Modality-specific Models:}
    \audiomod{} Qwen2-Audio-7B-Instruct \citep{chu_qwen2-audio_2024};
    \videomod{} Qwen2.5-VL-7B-Instruct \citep{bai_qwen25-vl_2025}, Qwen2-VL-7B-Instruct \citep{wang_qwen2-vl_2024}.
    \item \textbf{\multimodal{} Multimodal LLMs:} Phi-4-Multimodal-Instruct \citep{microsoft_phi-4-mini_2025}, Qwen2.5-Omni-7B \citep{xu_qwen25-omni_2025}, and Gemini-2.5-Flash \citep{comanici_gemini_2025}.
\end{itemize}

\subsection{Results}
\label{sec:results}
In the following, we report the results on \textsc{MuSaG}, for single and multimodal scenarios.

\begin{table*}[ht]
    \centering
    \resizebox{\linewidth}{!}{%
    \begin{tabular}{llccccccc}
    \toprule
    \textbf{Modality} & \textbf{Model} & \multicolumn{2}{c}{\textbf{Precision}} & \multicolumn{2}{c}{\textbf{Recall}} & \multicolumn{2}{c}{\textbf{F1}} & \multicolumn{1}{c}{\textbf{$\kappa$}}\\
    && \textsc{FullAgr.} & $\Delta$\textsc{Stand.} & \textsc{FullAgr.} & $\Delta$\textsc{Stand.} & \textsc{FullAgr.} & $\Delta$\textsc{Stand.} & \\
    \midrule
    N/A & random baseline & 49.34 &&49.31  && 49.19 && --\\
    \midrule
    \multirow{2}{*}{text \textmod{}} 
    & Qwen3-8B & 87.55 &{\color{blue} -4.23} & 88.01 &{\color{blue} -4.77} & 87.76 &{\color{blue} -4.48} & -- \\
    & human & 85.88 &{\color{blue} -3.67} & 86.55 &{\color{blue} -4.07} & 86.14 &{\color{blue} -3.84} & 53.13 \\
    \midrule
    \multirow{2}{*}{audio \audiomod{}}      
    & Gemini-2.5-flash & 76.82 &{\color{blue} +2.19} & 73.44 &{\color{blue} -1.77} & 66.83 &{\color{blue} +0.12} & --\\ 
    & human & \textbf{88.35} &{\color{blue} -4.72} &\textbf{ 87.62} &{\color{blue} -3.90} & \textbf{87.93} &{\color{blue} -4.21} & 68.01  \\
    \midrule
    \multirow{2}{*}{video \videomod{}} 
    & Gemini-2.5-flash & 59.34 &{\color{blue} +1.25 } & 59.87 &{\color{blue} + 0.87} & 59.1 &{\color{blue} +1.43} & -- \\
    & human & 69.12 &{\color{blue} -3.52} & 68.24 &{\color{blue} -4.13} & 68.55 &{\color{blue} -4.46} & 31.20  \\
    \midrule
     \multirow{2}{*}{audio-video \audiomod\videomod{}} & Gemini-2.5-flash &  79.44 & {\color{blue} -4.57} & 79.8 & {\color{blue} -4.88} & 79.61 &  {\color{blue} -4.72} \\
    & human &  100.00 & -- & 100.00 & -- & 100.00 & --& 100   \\
    \bottomrule
    \end{tabular}%
    }
    \caption{Results on \textsc{MuSaG-FullAgree}, the subset with full human agreement. 
We compare the best-performing models for each modality according to \cref{tab:results} against their corresponding results on \textsc{MuSaG} and human single-modality annotations. 
Human audio–video annotations are treated as the gold standard, reflecting how people naturally integrate multimodal cues in communication. $\Delta$\textit{Stand.} indicates the difference \textsc{MuSaG-FullAgree} - \textsc{MuSaG}.}
    \label{tab:sd100_and_human}
\end{table*}


%

\subsubsection{Unimodal Performance}
\label{subsec:automatic_evaluation}
We first evaluate each model using a single modality to understand how well sarcasm can be detected from transcript, audio, or video alone in \cref{tab:results}.
\paragraph{Text modality.}
Among the text-only LLMs, Qwen3-8B achieves the best results with an $83.28$ F1, outperforming smaller and older versions. Multimodal models evaluated on text alone generally perform slightly worse than dedicated text LLMs.

\paragraph{Audio modality.}
For audio-only input, unimodal audio LLMs show modest performance, with Qwen2-Audio-7B-Instruct achieving $55.18$ F1. Interestingly, the multimodal models Qwen2.5-Omni-7B and Gemini-2.5-flash outperform the audio-specific model: Gemini-2.5-flash reaches an F1-score of $66.95$, but only Qwen2.5-Omni-7B is able to leverage prosodic features to detect sarcasm and improve over its text-only performance.

\paragraph{Video modality.}
Sarcasm detection from video alone is particularly challenging. Vision-only models achieve moderate performance ($56.43$ F1), while multimodal models show mixed results. Gemini-2.5-flash again performs best ($60.53$ F1), whereas Phi-4-Multimodal-Instruct performs worse than chance.

\subsubsection{Multimodal Performance}
\label{subsec:multimodal_results}
We next examine model performance when multiple modalities are available, including combinations of text, audio, and video.

\paragraph{Text–audio and text–video.}
Combining transcripts with audio or video improves performance for most models. The commercial Gemini 2.5 Flash model achieves the highest scores for both text–audio ($86.91$ F1) and text–video ($83.63$ F1) inputs, showing a clear benefit from multimodal integration compared to single-modality settings. 
This improvement, however, is not consistent across all models. For instance, Qwen2.5-Omni-7B benefits from combining text with video, but not with audio.

\paragraph{Audio-video.}
When only audio and video are available, performance decreases relative to transcript-inclusive configurations, but still exceeds that of single-modality (audio-only or video-only) setups. Gemini-2.5-Flash achieves the best F1-score ($74.89$). This not only confirms that the transcript remains the most informative modality, but also highlights that in conditions resembling real human communication, where information is conveyed through speech and visual cues, commercial models still outperform open-source alternatives.

\paragraph{Text-audio-video.}
The full multimodal condition, including transcript, audio, and video, provides strong, but surprisingly not the best performance. Gemini-2.5-Flash achieves an F1-score of $83.38$, slightly lower than its transcript–audio performance (F1-score of $86.91$). We hypothesize that the addition of video can sometimes introduce noise or distract from the most informative cues. In contrast, Qwen2.5-Omni-7B benefits from combining all three modalities, achieving an F1 of $72.79$. These results suggest that transcript and audio carry the majority of sarcasm-relevant information, while video cues may only provide marginal gains or, for some models, slightly reduce performance.

\paragraph{Best performing open models.}
Among all evaluated systems, Gemini 2.5 Flash achieves the highest overall performance across all modalities except text-only input. Focusing on open-source models, Qwen2.5-Omni-7B consistently outperforms all other open models, showing strong results especially on audio-only input. Adding video or text to audio does not further improve its performance, only the full multimodal configuration achieves the best results overall.

\begin{table*}[ht]
    \centering
    \resizebox{\linewidth}{!}{%
    \begin{tabular}{llcccccc}
    \toprule
    \textbf{Modality} & \textbf{Model} & \multicolumn{2}{c}{\textbf{Precision}} & \multicolumn{2}{c}{\textbf{Recall}} & \multicolumn{2}{c}{\textbf{F1}}\\
    && Context & $\Delta$Stand. & Context. & $\Delta$Stand. & Context & $\Delta$Stand.  \\
    \midrule
    N/A & random baseline & 52.15 & -- & 52.15 & -- & 52.15 & -- \\
    \midrule
    \multirow{1}{*}{text \textmod{}} 
    & Qwen3-8B & 45.61 &{\color{blue} +37.71} & 45.77 &{\color{blue} +37.47} & 45.57 &{\color{blue} +37.71} \\
    \midrule
    \multirow{1}{*}{audio \audiomod{}}      
    & Gemini-2.5-flash & 55.31 & {\color{blue} +23.7} & 54.99 & {\color{blue} +16.68} & 53.01 & {\color{blue} +13.94} \\ 
    \midrule
    \multirow{1}{*}{video \videomod{}} 
    & Gemini-2.5-flash & 53.48 & {\color{blue} +7.11} & 52.70 & {\color{blue} +8.04} & 51.10 & {\color{blue} +9.43} \\
    \midrule
     \multirow{1}{*}{audio-video \audiomod\videomod{}} & Gemini-2.5-flash &  58.35 & {\color{blue} +16.52} & 55.15 & {\color{blue} +19.77} & 52.2 &  {\color{blue} +22.69} \\
    \bottomrule
    \end{tabular}%
    }
    \caption{Results on \textsc{MuSaG} with 15 seconds of extended context (\textit{Context}) in comparison to statements without context (\textit{Stand.}) for the best performing models for each modality according to \cref{tab:results}. $\Delta$\textit{Stand.} indicates the difference $Stand.-Context$. None of the models benefits from additional context.}
    \label{tab:results_ext_context}
\end{table*}
\subsection{Comparison with Human Classification} 

We now assess how model predictions align with human classifications, using \textsc{MuSaG-FullAgree} as a gold standard. This subset contains only examples with full annotator agreement, reflecting how people naturally perceive sarcasm in multimodal communication. We evaluate the best-performing model per modality, Qwen3-8B for text, and Gemini-2.5-flash for audio, video, and audio-video, against human classifications based on the corresponding single modality (transcript, audio, or video only).

\paragraph{Representative Examples.} Disagreements between humans and models arise in both directions: models sometimes predict sarcasm where humans do not, and vice versa. Representative examples of such disagreements across all modalities are shown in \cref{tab:disagreements_modality}.

\paragraph{Humans vs.\ Models.} 
\Cref{tab:sd100_and_human} compares human and model performance on \textsc{MuSaG-FullAgree} across all modalities, where humans and models were asked to classify sarcasm based solely on a single modality (text, audio, or video).

Humans derive the most reliable cues from audio ($87.93$~F1), suggesting that prosodic features such as tone and intonation provide strong indicators of sarcasm. In contrast, multimodal models are not yet able to leverage these cues effectively: humans outperform models by substantial margins, nearly $21$~F1 points for audio and $10$~F1 points for video. For video-only input, human performance drops to $68.6$~F1, indicating that visual cues alone are often insufficient for sarcasm detection, with model performance similarly decreasing to $58.5$~F1.

Generally, most models perform better on \textsc{MuSaG-FullAgree} than on the full dataset (indicated in blue in \cref{tab:sd100_and_human}), reflecting that examples with perfect human agreement are likely less ambiguous and thus easier to interpret.


Results on \textsc{MuSaG-FullAgree} for all models, not only the best performing ones, can be found in \cref{tab:results_sd_100} in \cref{app:results_sd_100}.

\subsection{Extending the Context}
\label{subsec:extended_context}

Lastly, we investigate whether providing models with additional temporal context improves sarcasm detection performance, for the best model for each modality according to \cref{tab:results}.
Specifically, we extend the input by 15 seconds surrounding the target utterance, while explicitly including the sentence transcript to be classified in the model prompt. We report these results in \cref{tab:results_ext_context}.

Surprisingly, this makes the task significantly harder for all models, and performance drops to chance. We hypothesize that the models, even when explicitly provided with the target utterance, may struggle to attribute their decision to the correct segment within the extended input. The added temporal context likely introduces distracting or conflicting cues, making it harder for the model to focus on the relevant part of the provided signal.

Qualitative examples illustrating these three outcomes, 1) cases where context helped, 2) hurt, or 3) had no effect on classification, are shown in Table~\ref{tab:context_effects} in Appendix~\ref{app:examples}. Notably, the \textit{Degraded} example in Table~\ref{tab:context_effects} suggests that surrounding utterances can introduce misleading cues: the phrase \textit{„Das klingt ja erstmal so, als wäre jetzt alles in trockenen Tüchern"} adopts a tone that superficially resembles sarcasm, apparently causing three out of four models to misclassify the subsequent non-sarcastic target sentence.

This finding has important implications for real-world deployment. In natural conversations, utterances never occur in isolation, they are embedded in a broader discourse with preceding and following context. If models are already misled by a mere 15 seconds of surrounding signal, their reliability in practical applications such as opinion mining, content moderation, or social media analysis may be substantially lower than isolated-utterance benchmarks suggest. Addressing this sensitivity to context should therefore be a priority for future work on sarcasm detection in the wild.

\section{Conclusion}
We introduced \textsc{MuSaG}, the first manually curated German multimodal sarcasm dataset with independent annotations for text, audio, and video modalities. The dataset includes both full statements and modality-specific labels, enabling fine-grained analysis of multimodal sarcasm understanding. 
Moreover, we also release  \textsc{MuSaG-FullAgree}, a subset with full annotator agreement (annotated in audio–video), which can serve as a gold standard for how humans perceive sarcasm and can be used to evaluate how well humans and models perform when only partial modalities are available.
\textsc{MuSaG} will be released publicly to support research on multimodal language models.

Our benchmarking experiments show that humans rely primarily on audio cues, followed by text and then video, indicating that the strongest signals for sarcasm lie in prosody and intonation. In contrast, models perform strongest on text, revealing that they fail to fully exploit audio cues and are not yet capable of genuine multimodal understanding. While commercial models generally outperform open-source models, all architectures struggle to integrate non-textual information effectively. 

These findings underscore the challenges of building systems for nuanced sarcasm detection, highlighting that current multimodal models are still unable to effectively leverage non-textual cues, emphasizing the value of \textsc{MuSaG} as a benchmark for developing and evaluating truly multimodal models.

\section{Ethical Considerations}
All annotators participated voluntarily and will be acknowledged by name upon paper acceptance. The dataset only includes publicly available content, and we release links to the original videos rather than the video files themselves. Researchers should note that sarcasm detection can reflect cultural and subjective biases.

\section{Acknowledgements}
Part of this work received support from the European Union’s Horizon research and innovation programme under grant agreement No 101135798, project Meetween (My Personal AI Mediator for Virtual MEETtings BetWEEN People).
\nocite{*}
\section{Bibliographical References}\label{sec:reference}

\bibliographystyle{lrec2026-natbib}
\bibliography{research}

\begin{thebibliography}{16}
\expandafter\ifx\csname natexlab\endcsname\relax\def\natexlab#1{#1}\fi

\bibitem[{Alnajjar and Hämäläinen(2021)}]{alnajjar_que_2021}
Alnajjar, Khalid and Hämäläinen, Mika. 2021.
\newblock \href {https://doi.org/10.18653/v1/2021.maiworkshop-1.9} {\emph{¡{Qué} maravilla! {Multimodal} {Sarcasm} {Detection} in {Spanish}: a {Dataset} and a {Baseline}}}.
\newblock Association for Computational Linguistics.
\newblock PID \href{https://zenodo.org/records/4701383}{https://zenodo.org/records/4701383}.

\bibitem[{Bedi et~al.(2023)Bedi, Kumar, Akhtar, and Chakraborty}]{bedi_multi-modal_2023}
Bedi, Manjot and Kumar, Shivani and Akhtar, Md Shad and Chakraborty, Tanmoy. 2023.
\newblock \href {https://doi.org/10.1109/TAFFC.2021.3083522} {\emph{Multi-{Modal} {Sarcasm} {Detection} and {Humor} {Classification} in {Code}-{Mixed} {Conversations}}}.
\newblock IEEE Transactions on Affective Computing.
\newblock PID \href{https://github.com/LCS2-IIITD/MSH-COMICS}{https://github.com/LCS2-IIITD/MSH-COMICS}.

\bibitem[{Cai et~al.(2019)Cai, Cai, and Wan}]{cai_multi-modal_2019}
Cai, Yitao and Cai, Huiyu and Wan, Xiaojun. 2019.
\newblock \href {https://doi.org/10.18653/v1/P19-1239} {\emph{Multi-{Modal} {Sarcasm} {Detection} in {Twitter} with {Hierarchical} {Fusion} {Model}}}.
\newblock Association for Computational Linguistics.
\newblock PID \href{https://github.com/headacheboy/data-of-multimodal-sarcasm-detection}{https://github.com/headacheboy/data-of-multimodal-sarcasm-detection}.

\bibitem[{Castro et~al.(2019)Castro, Hazarika, Pérez-Rosas, Zimmermann, Mihalcea, and Poria}]{castro_towards_2019}
Castro, Santiago and Hazarika, Devamanyu and Pérez-Rosas, Verónica and Zimmermann, Roger and Mihalcea, Rada and Poria, Soujanya. 2019.
\newblock \href {https://doi.org/10.18653/v1/P19-1455} {\emph{Towards {Multimodal} {Sarcasm} {Detection} ({An} \_Obviously\_ {Perfect} {Paper})}}.
\newblock Association for Computational Linguistics.
\newblock PID \href{https://github.com/soujanyaporia/MUStARD}{https://github.com/soujanyaporia/MUStARD}.

\bibitem[{Davidov et~al.(2010)Davidov, Tsur, and Rappoport}]{davidov_semi-supervised_2010}
Davidov, Dmitry and Tsur, Oren and Rappoport, Ari. 2010.
\newblock \href {https://aclanthology.org/W10-2914/} {\emph{Semi-{Supervised} {Recognition} of {Sarcasm} in {Twitter} and {Amazon}}}.
\newblock Association for Computational Linguistics.

\bibitem[{González-Ibáñez et~al.(2011)González-Ibáñez, Muresan, and Wacholder}]{gonzalez-ibanez_identifying_2011}
González-Ibáñez, Roberto and Muresan, Smaranda and Wacholder, Nina. 2011.
\newblock \href {https://aclanthology.org/P11-2102/} {\emph{Identifying {Sarcasm} in {Twitter}: {A} {Closer} {Look}}}.
\newblock Association for Computational Linguistics.

\bibitem[{Joshi et~al.(2016)Joshi, Tripathi, Bhattacharyya, and Carman}]{joshi_harnessing_2016}
Joshi, Aditya and Tripathi, Vaibhav and Bhattacharyya, Pushpak and Carman, Mark J. 2016.
\newblock \href {https://doi.org/10.18653/v1/K16-1015} {\emph{Harnessing {Sequence} {Labeling} for {Sarcasm} {Detection} in {Dialogue} from {TV} {Series} ‘{Friends}'}}.
\newblock Association for Computational Linguistics.

\bibitem[{Qin et~al.(2023)Qin, Huang, Chen, Cai, Zhang, Liang, Che, and Xu}]{qin-etal-2023-mmsd2}
Qin, Libo and Huang, Shijue and Chen, Qiguang and Cai, Chenran and Zhang, Yudi and Liang, Bin and Che, Wanxiang and Xu, Ruifeng. 2023.
\newblock \href {https://doi.org/10.18653/v1/2023.findings-acl.689} {\emph{{MMSD}2.0: Towards a Reliable Multi-modal Sarcasm Detection System}}.
\newblock Association for Computational Linguistics.
\newblock PID \href{https://github.com/JoeYing1019/MMSD2.0}{https://github.com/JoeYing1019/MMSD2.0}.

\bibitem[{Ray et~al.(2022)Ray, Mishra, Nunna, and Bhattacharyya}]{ray_multimodal_2022}
Ray, Anupama and Mishra, Shubham and Nunna, Apoorva and Bhattacharyya, Pushpak. 2022.
\newblock \href {https://aclanthology.org/2022.lrec-1.756/} {\emph{A {Multimodal} {Corpus} for {Emotion} {Recognition} in {Sarcasm}}}.
\newblock European Language Resources Association.

\bibitem[{Sangwan et~al.(2020)Sangwan, Akhtar, Behera, and Ekbal}]{sangwan_i_2020}
Sangwan, Suyash and Akhtar, Md Shad and Behera, Pranati and Ekbal, Asif. 2020.
\newblock \href {https://doi.org/10.1109/IJCNN48605.2020.9206905} {\emph{I didn’t mean what {I} wrote! {Exploring} {Multimodality} for {Sarcasm} {Detection}}}.
\newblock 2020 {International} {Joint} {Conference} on {Neural} {Networks} ({IJCNN}).
\newblock PID \href{http://www.iitp.ac.in/ai-nlp-ml/resources.htm}{http://www.iitp.ac.in/ai-nlp-ml/resources.htm}.
\newblock ISSN: 2161-4407.

\bibitem[{Schifanella et~al.(2016)Schifanella, de~Juan, Tetreault, and Cao}]{schifanella_detecting_2016}
Schifanella, Rossano and de Juan, Paloma and Tetreault, Joel and Cao, LiangLiang. 2016.
\newblock \href {https://doi.org/10.1145/2964284.2964321} {\emph{Detecting {Sarcasm} in {Multimodal} {Social} {Platforms}}}.
\newblock Association for Computing Machinery, {MM} '16.

\bibitem[{Tepperman et~al.(2006)Tepperman, Traum, and Narayanan}]{tepperman_yeah_2006}
Tepperman, Joseph and Traum, David and Narayanan, Shrikanth. 2006.
\newblock \href {https://doi.org/10.21437/interspeech.2006-507} {\emph{yeah right: sarcasm recognition for spoken dialogue systems}}.
\newblock ISCA.

\bibitem[{Tsur et~al.(2010)Tsur, Davidov, and Rappoport}]{tsur_icwsm_2010}
Tsur, Oren and Davidov, Dmitry and Rappoport, Ari. 2010.
\newblock \href {https://doi.org/10.1609/icwsm.v4i1.14018} {\emph{{ICWSM} — {A} {Great} {Catchy} {Name}: {Semi}-{Supervised} {Recognition} of {Sarcastic} {Sentences} in {Online} {Product} {Reviews}}}.
\newblock Proceedings of the International AAAI Conference on Web and Social Media.

\bibitem[{Wallace et~al.(2014)Wallace, Choe, Kertz, and Charniak}]{wallace_humans_2014}
Wallace, Byron C. and Choe, Do Kook and Kertz, Laura and Charniak, Eugene. 2014.
\newblock \href {https://doi.org/10.3115/v1/P14-2084} {\emph{Humans {Require} {Context} to {Infer} {Ironic} {Intent} (so {Computers} {Probably} do, too)}}.
\newblock Association for Computational Linguistics.
\newblock PID \href{https://github.com/bwallace/ACL-2014-irony}{https://github.com/bwallace/ACL-2014-irony}.

\bibitem[{Yue et~al.(2024)Yue, Shi, Mao, Hu, and Cambria}]{yue_sarcnet_2024}
Yue, Tan and Shi, Xuzhao and Mao, Rui and Hu, Zonghai and Cambria, Erik. 2024.
\newblock \href {https://aclanthology.org/2024.lrec-main.1248/} {\emph{{SarcNet}: {A} {Multilingual} {Multimodal} {Sarcasm} {Detection} {Dataset}}}.
\newblock ELRA and ICCL.
\newblock PID \href{https://github.com/yuetanbupt/SarcNet}{https://github.com/yuetanbupt/SarcNet}.

\bibitem[{Zhang et~al.(2023)Zhang, Yu, Guo, Wang, Zhao, Uprety, Song, Li, and Qin}]{zhang_cmma_2023}
Zhang, Yazhou and Yu, Yang and Guo, Qing and Wang, Benyou and Zhao, Dongming and Uprety, Sagar and Song, Dawei and Li, Qiuchi and Qin, Jing. 2023.
\newblock \emph{{CMMA}: benchmarking multi-affection detection in chinese multi-modal conversations}.
\newblock Curran Associates Inc., {NIPS} '23.
\newblock PID \href{https://github.com/annoymity2022/Chinese-Dataset}{https://github.com/annoymity2022/Chinese-Dataset}.

\end{thebibliography}


\begin{thebibliography}{17}
\expandafter\ifx\csname natexlab\endcsname\relax\def\natexlab#1{#1}\fi

\bibitem[{Abouelenin et~al.(2025)Abouelenin, Ashfaq, Atkinson, Awadalla, Bach, Bao, Benhaim, Cai, Chaudhary, Chen, Chen, Chen, Chen, Chen, Chen, Chen, Dai, Dai, Fan, Gao, Gao, Garg, Goswami, Hao, Hendy, Hu, Jin, Khademi, Kim, Kim, Lee, Li, Li, Liang, Lin, Lin, Liu, Liu, Lopez, Luo, Madan, Mazalov, Mitra, Mousavi, Nguyen, Pan, Perez-Becker, Platin, Portet, Qiu, Ren, Ren, Roy, Shang, Shen, Singhal, Som, Song, Sych, Vaddamanu, Wang, Wang, Wang, Wu, Xu, Xu, Yang, Yang, Yu, Zabir, Zhang, Zhang, Zhang, and Zhou}]{microsoft_phi-4-mini_2025}
Abdelrahman Abouelenin, Atabak Ashfaq, Adam Atkinson, Hany Awadalla, Nguyen Bach, Jianmin Bao, Alon Benhaim, Martin Cai, Vishrav Chaudhary, Congcong Chen, Dong Chen, Dongdong Chen, Junkun Chen, Weizhu Chen, Yen-Chun Chen, Yi-ling Chen, Qi~Dai, Xiyang Dai, Ruchao Fan, Mei Gao, Min Gao, Amit Garg, Abhishek Goswami, Junheng Hao, Amr Hendy, Yuxuan Hu, Xin Jin, Mahmoud Khademi, Dongwoo Kim, Young~Jin Kim, Gina Lee, Jinyu Li, Yunsheng Li, Chen Liang, Xihui Lin, Zeqi Lin, Mengchen Liu, Yang Liu, Gilsinia Lopez, Chong Luo, Piyush Madan, Vadim Mazalov, Arindam Mitra, Ali Mousavi, Anh Nguyen, Jing Pan, Daniel Perez-Becker, Jacob Platin, Thomas Portet, Kai Qiu, Bo~Ren, Liliang Ren, Sambuddha Roy, Ning Shang, Yelong Shen, Saksham Singhal, Subhojit Som, Xia Song, Tetyana Sych, Praneetha Vaddamanu, Shuohang Wang, Yiming Wang, Zhenghao Wang, Haibin Wu, Haoran Xu, Weijian Xu, Yifan Yang, Ziyi Yang, Donghan Yu, Ishmam Zabir, Jianwen Zhang, Li~Lyna Zhang, Yunan Zhang, and Xiren Zhou. 2025.
\newblock \href {https://doi.org/10.48550/arXiv.2503.01743} {Phi-4-{Mini} {Technical} {Report}: {Compact} yet {Powerful} {Multimodal} {Language} {Models} via {Mixture}-of-{LoRAs}}.
\newblock ArXiv:2503.01743 [cs].

\bibitem[{Bai et~al.(2025)Bai, Chen, Liu, Wang, Ge, Song, Dang, Wang, Wang, Tang, Zhong, Zhu, Yang, Li, Wan, Wang, Ding, Fu, Xu, Ye, Zhang, Xie, Cheng, Zhang, Yang, Xu, and Lin}]{bai_qwen25-vl_2025}
Shuai Bai, Keqin Chen, Xuejing Liu, Jialin Wang, Wenbin Ge, Sibo Song, Kai Dang, Peng Wang, Shijie Wang, Jun Tang, Humen Zhong, Yuanzhi Zhu, Mingkun Yang, Zhaohai Li, Jianqiang Wan, Pengfei Wang, Wei Ding, Zheren Fu, Yiheng Xu, Jiabo Ye, Xi~Zhang, Tianbao Xie, Zesen Cheng, Hang Zhang, Zhibo Yang, Haiyang Xu, and Junyang Lin. 2025.
\newblock \href {https://doi.org/10.48550/arXiv.2502.13923} {Qwen2.5-{VL} {Technical} {Report}}.
\newblock ArXiv:2502.13923 [cs].

\bibitem[{Chu et~al.(2024)Chu, Xu, Yang, Wei, Wei, Guo, Leng, Lv, He, Lin, Zhou, and Zhou}]{chu_qwen2-audio_2024}
Yunfei Chu, Jin Xu, Qian Yang, Haojie Wei, Xipin Wei, Zhifang Guo, Yichong Leng, Yuanjun Lv, Jinzheng He, Junyang Lin, Chang Zhou, and Jingren Zhou. 2024.
\newblock \href {https://doi.org/10.48550/arXiv.2407.10759} {Qwen2-{Audio} {Technical} {Report}}.
\newblock ArXiv:2407.10759 [eess].

\bibitem[{Comanici et~al.(2025)Comanici, Bieber, Schaekermann, Pasupat, Sachdeva, Dhillon, Blistein, Ram, Zhang, Rosen, Marris, Petulla, Gaffney, Aharoni, Lintz, Cardal~Pais, Jacobsson, Szpektor, Jiang, Haridasan, Omran, Saunshi, Bahri, Mishra, Chu, Boyd, Hekman, Parisi, Zhang, Kawintiranon, Bedrax-Weiss, Wang, Xu, Purkiss, Mendlovic, Deutel, Nguyen, Langley, Korn, Rossazza, Ramé, Waghmare, Miller, Byrd, Sheshan, Hadsell Sangnie~Bhardwaj, Janus, Rissa, Horgan, Silver, Wahid, Brin, Raimond, Kloboves, Wang, Gundavarapu, Shumailov, Wang, Pajarskas, Heyward, Nikoltchev, Kula, Zhou, Garrett, Kafle, Arik, Goel, Yang, Park, Kojima, Mahmoudieh, Kavukcuoglu, Chen, Fritz, Bulyenov, Roy, Paparas, Shemtov, Chen, Strudel, Reitter, Roy, Vlasov, Ryu, Leichner, Yang, Mariet, Vnukov, Sohn, Stuart, Liang, Chen, Rawlani, Koh, Co-Reyes, Lai, Banzal, Vytiniotis, Mei, Cai, Badawi, Fry, Hartman, Zheng, Jia, Keeling, Louis, Chen, Robles, Hung, Zhou, Saxena, Goenka, Ma, Fisher, Hazan~Taege, Graves, Steiner, Li, Nguyen, Sukthankar,
  Stanton, Eslami, Shen, Akin, Guseynov, Zhou, Alayrac, Joulin, Farkash, Thapliyal, Roller, Shazeer, Davchev, Koo, Forbes-Pollard, Audhkhasi, Farquhar, Mayrav~Gilady, Song, Aslanides, Mendolicchio, Parrish, Blitzer, Gupta, Ju, Yang, Datta, Tacchetti, Vaibhav~Mehta, Dibb, Gupta, Piccinini, Hadsell, Rajayogam, Jiang, Griffin, Sundberg, Hayes, Frolov, Xie, Zhang, Dasgupta, Kalra, Shani, Macherey, Huang, MacDermed, Duddu, Zacchello, Yang, Lo, Hui, Kastelic, Gasaway, Tan, Yue, Barrio, Wieting, Yang, Nystrom, Demmessie, Levskaya, Viola, Tekur, Billock, Necula, Joshi, Schaeffer, Lokhande, Sorokin, Shenoy, Chen, Collier, Li, Bos, Wichers, Lee, Pouget, and Thangaraj}]{comanici_gemini_2025}
Gheorghe Comanici, Eric Bieber, Mike Schaekermann, Ice Pasupat, Noveen Sachdeva, Inderjit Dhillon, Marcel Blistein, Ori Ram, Dan Zhang, Evan Rosen, Luke Marris, Sam Petulla, Colin Gaffney, Asaf Aharoni, Nathan Lintz, Tiago Cardal~Pais, Henrik Jacobsson, Idan Szpektor, Nan-Jiang Jiang, Krishna Haridasan, Ahmed Omran, Nikunj Saunshi, Dara Bahri, Gaurav Mishra, Eric Chu, Toby Boyd, Brad Hekman, Aaron Parisi, Chaoyi Zhang, Kornraphop Kawintiranon, Tania Bedrax-Weiss, Oliver Wang, Ya~Xu, Ollie Purkiss, Uri Mendlovic, Ilaï Deutel, Nam Nguyen, Adam Langley, Flip Korn, Lucia Rossazza, Alexandre Ramé, Sagar Waghmare, Helen Miller, Nathan Byrd, Ashrith Sheshan, Raia Hadsell Sangnie~Bhardwaj, Pawel Janus, Tero Rissa, Dan Horgan, Sharon Silver, Ayzaan Wahid, Sergey Brin, Yves Raimond, Klemen Kloboves, Cindy Wang, Nitesh~Bharadwaj Gundavarapu, Ilia Shumailov, Bo~Wang, Mantas Pajarskas, Joe Heyward, Martin Nikoltchev, Maciej Kula, Hao Zhou, Zachary Garrett, Sushant Kafle, Sercan Arik, Ankita Goel, Mingyao Yang, Jiho
  Park, Koji Kojima, Parsa Mahmoudieh, Koray Kavukcuoglu, Grace Chen, Doug Fritz, Anton Bulyenov, Sudeshna Roy, Dimitris Paparas, Hadar Shemtov, Bo-Juen Chen, Robin Strudel, David Reitter, Aurko Roy, Andrey Vlasov, Changwan Ryu, Chas Leichner, Haichuan Yang, Zelda Mariet, Denis Vnukov, Tim Sohn, Amy Stuart, Wei Liang, Minmin Chen, Praynaa Rawlani, Christy Koh, JD~Co-Reyes, Guangda Lai, Praseem Banzal, Dimitrios Vytiniotis, Jieru Mei, Mu~Cai, Mohammed Badawi, Corey Fry, Ale Hartman, Daniel Zheng, Eric Jia, James Keeling, Annie Louis, Ying Chen, Efren Robles, Wei-Chih Hung, Howard Zhou, Nikita Saxena, Sonam Goenka, Olivia Ma, Zach Fisher, Mor Hazan~Taege, Emily Graves, David Steiner, Yujia Li, Sarah Nguyen, Rahul Sukthankar, Joe Stanton, Ali Eslami, Gloria Shen, Berkin Akin, Alexey Guseynov, Yiqian Zhou, Jean-Baptiste Alayrac, Armand Joulin, Efrat Farkash, Ashish Thapliyal, Stephen Roller, Noam Shazeer, Todor Davchev, Terry Koo, Hannah Forbes-Pollard, Kartik Audhkhasi, Greg Farquhar, Adi Mayrav~Gilady, Maggie
  Song, John Aslanides, Piermaria Mendolicchio, Alicia Parrish, John Blitzer, Pramod Gupta, Xiaoen Ju, Xiaochen Yang, Puranjay Datta, Andrea Tacchetti, Sanket Vaibhav~Mehta, Gregory Dibb, Shubham Gupta, Federico Piccinini, Raia Hadsell, Sujee Rajayogam, Jiepu Jiang, Patrick Griffin, Patrik Sundberg, Jamie Hayes, Alexey Frolov, Tian Xie, Adam Zhang, Kingshuk Dasgupta, Uday Kalra, Lior Shani, Klaus Macherey, Tzu-Kuo Huang, Liam MacDermed, Karthik Duddu, Paulo Zacchello, Zi~Yang, Jessica Lo, Kai Hui, Matej Kastelic, Derek Gasaway, Qijun Tan, Summer Yue, Pablo Barrio, John Wieting, Weel Yang, Andrew Nystrom, Solomon Demmessie, Anselm Levskaya, Fabio Viola, Chetan Tekur, Greg Billock, George Necula, Mandar Joshi, Rylan Schaeffer, Swachhand Lokhande, Christina Sorokin, Pradeep Shenoy, Mia Chen, Mark Collier, Hongji Li, Taylor Bos, Nevan Wichers, Sun~Jae Lee, Angéline Pouget, and Santhosh Thangaraj. 2025.
\newblock \href {https://doi.org/10.48550/arXiv.2507.06261} {Gemini 2.5: {Pushing} the {Frontier} with {Advanced} {Reasoning}, {Multimodality}, {Long} {Context}, and {Next} {Generation} {Agentic} {Capabilities}}.
\newblock ADS Bibcode: 2025arXiv250706261C.

\bibitem[{Farabi et~al.(2024)Farabi, Ranasinghe, Kanojia, Kong, and Zampieri}]{ijcai2024p887}
Shafkat Farabi, Tharindu Ranasinghe, Diptesh Kanojia, Yu~Kong, and Marcos Zampieri. 2024.
\newblock \href {https://doi.org/10.24963/ijcai.2024/887} {A survey of multimodal sarcasm detection}.
\newblock In \emph{Proceedings of the Thirty-Third International Joint Conference on Artificial Intelligence, {IJCAI-24}}, pages 8020--8028. International Joint Conferences on Artificial Intelligence Organization.
\newblock Survey Track.

\bibitem[{Frenda(2018)}]{role_of_sarcasm}
Simona Frenda. 2018.
\newblock The role of sarcasm in hate speech. a multilingual perspective.
\newblock In \emph{Proceedings of the Doctoral Symposium of the XXXIVInternational Conference of the Spanish Society for Natural Language Processing (SEPLN 2018)}, volume Vol-2251.

\bibitem[{Joshi et~al.(2017)Joshi, Bhattacharyya, and Carman}]{joshi-sarcasm-survey}
Aditya Joshi, Pushpak Bhattacharyya, and Mark~J. Carman. 2017.
\newblock \href {https://doi.org/10.1145/3124420} {Automatic sarcasm detection: A survey}.
\newblock \emph{ACM Comput. Surv.}, 50(5).

\bibitem[{Landis and Koch(1977)}]{landis_measurement_1977}
J.~Richard Landis and Gary~G. Koch. 1977.
\newblock \href {https://doi.org/10.2307/2529310} {The {Measurement} of {Observer} {Agreement} for {Categorical} {Data}}.
\newblock \emph{Biometrics}, 33(1):159--174.
\newblock Publisher: International Biometric Society.

\bibitem[{Liu et~al.(2025)Liu, Xu, Wu, Yuan, Yang, Zhou, Liu, Guan, Wang, Yu, McAuley, Ai, and Huang}]{DBLP:conf/naacl/LiuXWYYZLGWYMAH25}
Xiaoyu Liu, Paiheng Xu, Junda Wu, Jiaxin Yuan, Yifan Yang, Yuhang Zhou, Fuxiao Liu, Tianrui Guan, Haoliang Wang, Tong Yu, Julian~J. McAuley, Wei Ai, and Furong Huang. 2025.
\newblock \href {https://doi.org/10.18653/V1/2025.FINDINGS-NAACL.427} {Large language models and causal inference in collaboration: {A} comprehensive survey}.
\newblock In \emph{Findings of the Association for Computational Linguistics: {NAACL} 2025, Albuquerque, New Mexico, USA, April 29 - May 4, 2025}, pages 7668--7684. Association for Computational Linguistics.

\bibitem[{Maynard and Greenwood(2014)}]{maynard-greenwood-2014-cares}
Diana Maynard and Mark Greenwood. 2014.
\newblock \href {https://aclanthology.org/L14-1527/} {Who cares about sarcastic tweets? investigating the impact of sarcasm on sentiment analysis.}
\newblock In \emph{Proceedings of the Ninth International Conference on Language Resources and Evaluation ({LREC}'14)}, pages 4238--4243, Reykjavik, Iceland. European Language Resources Association (ELRA).

\bibitem[{Pan et~al.(2020)Pan, Lin, Fu, Qi, and Wang}]{pan_modeling_2020}
Hongliang Pan, Zheng Lin, Peng Fu, Yatao Qi, and Weiping Wang. 2020.
\newblock \href {https://doi.org/10.18653/v1/2020.findings-emnlp.124} {Modeling {Intra} and {Inter}-modality {Incongruity} for {Multi}-{Modal} {Sarcasm} {Detection}}.
\newblock In \emph{Findings of the {Association} for {Computational} {Linguistics}: {EMNLP} 2020}, pages 1383--1392, Online. Association for Computational Linguistics.

\bibitem[{Qwen et~al.(2025)Qwen, Yang, Yang, Zhang, Hui, Zheng, Yu, Li, Liu, Huang, Wei, Lin, Yang, Tu, Zhang, Yang, Yang, Zhou, Lin, Dang, Lu, Bao, Yang, Yu, Li, Xue, Zhang, Zhu, Men, Lin, Li, Tang, Xia, Ren, Ren, Fan, Su, Zhang, Wan, Liu, Cui, Zhang, and Qiu}]{qwen_qwen25_2025}
Qwen, An~Yang, Baosong Yang, Beichen Zhang, Binyuan Hui, Bo~Zheng, Bowen Yu, Chengyuan Li, Dayiheng Liu, Fei Huang, Haoran Wei, Huan Lin, Jian Yang, Jianhong Tu, Jianwei Zhang, Jianxin Yang, Jiaxi Yang, Jingren Zhou, Junyang Lin, Kai Dang, Keming Lu, Keqin Bao, Kexin Yang, Le~Yu, Mei Li, Mingfeng Xue, Pei Zhang, Qin Zhu, Rui Men, Runji Lin, Tianhao Li, Tianyi Tang, Tingyu Xia, Xingzhang Ren, Xuancheng Ren, Yang Fan, Yang Su, Yichang Zhang, Yu~Wan, Yuqiong Liu, Zeyu Cui, Zhenru Zhang, and Zihan Qiu. 2025.
\newblock \href {https://doi.org/10.48550/arXiv.2412.15115} {Qwen2.5 {Technical} {Report}}.
\newblock ArXiv:2412.15115 [cs].

\bibitem[{Radford et~al.(2022)Radford, Kim, Xu, Brockman, McLeavey, and Sutskever}]{radford2022whisper}
Alec Radford, Jong~Wook Kim, Tao Xu, Greg Brockman, Christine McLeavey, and Ilya Sutskever. 2022.
\newblock \href {https://doi.org/10.48550/ARXIV.2212.04356} {Robust speech recognition via large-scale weak supervision}.

\bibitem[{Wang et~al.(2024)Wang, Bai, Tan, Wang, Fan, Bai, Chen, Liu, Wang, Ge, Fan, Dang, Du, Ren, Men, Liu, Zhou, Zhou, and Lin}]{wang_qwen2-vl_2024}
Peng Wang, Shuai Bai, Sinan Tan, Shijie Wang, Zhihao Fan, Jinze Bai, Keqin Chen, Xuejing Liu, Jialin Wang, Wenbin Ge, Yang Fan, Kai Dang, Mengfei Du, Xuancheng Ren, Rui Men, Dayiheng Liu, Chang Zhou, Jingren Zhou, and Junyang Lin. 2024.
\newblock \href {https://doi.org/10.48550/arXiv.2409.12191} {Qwen2-{VL}: {Enhancing} {Vision}-{Language} {Model}'s {Perception} of the {World} at {Any} {Resolution}}.
\newblock ArXiv:2409.12191 [cs].

\bibitem[{Xu et~al.(2025)Xu, Guo, He, Hu, He, Bai, Chen, Wang, Fan, Dang, Zhang, Wang, Chu, and Lin}]{xu_qwen25-omni_2025}
Jin Xu, Zhifang Guo, Jinzheng He, Hangrui Hu, Ting He, Shuai Bai, Keqin Chen, Jialin Wang, Yang Fan, Kai Dang, Bin Zhang, Xiong Wang, Yunfei Chu, and Junyang Lin. 2025.
\newblock \href {https://doi.org/10.48550/arXiv.2503.20215} {Qwen2.5-{Omni} {Technical} {Report}}.
\newblock ArXiv:2503.20215 [cs].

\bibitem[{Yang et~al.(2025)Yang, Li, Yang, Zhang, Hui, Zheng, Yu, Gao, Huang, Lv, Zheng, Liu, Zhou, Huang, Hu, Ge, Wei, Lin, Tang, Yang, Tu, Zhang, Yang, Yang, Zhou, Zhou, Lin, Dang, Bao, Yang, Yu, Deng, Li, Xue, Li, Zhang, Wang, Zhu, Men, Gao, Liu, Luo, Li, Tang, Yin, Ren, Wang, Zhang, Ren, Fan, Su, Zhang, Zhang, Wan, Liu, Wang, Cui, Zhang, Zhou, and Qiu}]{yang_qwen3_2025}
An~Yang, Anfeng Li, Baosong Yang, Beichen Zhang, Binyuan Hui, Bo~Zheng, Bowen Yu, Chang Gao, Chengen Huang, Chenxu Lv, Chujie Zheng, Dayiheng Liu, Fan Zhou, Fei Huang, Feng Hu, Hao Ge, Haoran Wei, Huan Lin, Jialong Tang, Jian Yang, Jianhong Tu, Jianwei Zhang, Jianxin Yang, Jiaxi Yang, Jing Zhou, Jingren Zhou, Junyang Lin, Kai Dang, Keqin Bao, Kexin Yang, Le~Yu, Lianghao Deng, Mei Li, Mingfeng Xue, Mingze Li, Pei Zhang, Peng Wang, Qin Zhu, Rui Men, Ruize Gao, Shixuan Liu, Shuang Luo, Tianhao Li, Tianyi Tang, Wenbiao Yin, Xingzhang Ren, Xinyu Wang, Xinyu Zhang, Xuancheng Ren, Yang Fan, Yang Su, Yichang Zhang, Yinger Zhang, Yu~Wan, Yuqiong Liu, Zekun Wang, Zeyu Cui, Zhenru Zhang, Zhipeng Zhou, and Zihan Qiu. 2025.
\newblock \href {https://doi.org/10.48550/arXiv.2505.09388} {Qwen3 {Technical} {Report}}.
\newblock ArXiv:2505.09388 [cs].

\bibitem[{Yang et~al.(2024)Yang, Yang, Hui, Zheng, Yu, Zhou, Li, Li, Liu, Huang, Dong, Wei, Lin, Tang, Wang, Yang, Tu, Zhang, Ma, Yang, Xu, Zhou, Bai, He, Lin, Dang, Lu, Chen, Yang, Li, Xue, Ni, Zhang, Wang, Peng, Men, Gao, Lin, Wang, Bai, Tan, Zhu, Li, Liu, Ge, Deng, Zhou, Ren, Zhang, Wei, Ren, Liu, Fan, Yao, Zhang, Wan, Chu, Liu, Cui, Zhang, Guo, and Fan}]{yang_qwen2_2024}
An~Yang, Baosong Yang, Binyuan Hui, Bo~Zheng, Bowen Yu, Chang Zhou, Chengpeng Li, Chengyuan Li, Dayiheng Liu, Fei Huang, Guanting Dong, Haoran Wei, Huan Lin, Jialong Tang, Jialin Wang, Jian Yang, Jianhong Tu, Jianwei Zhang, Jianxin Ma, Jianxin Yang, Jin Xu, Jingren Zhou, Jinze Bai, Jinzheng He, Junyang Lin, Kai Dang, Keming Lu, Keqin Chen, Kexin Yang, Mei Li, Mingfeng Xue, Na~Ni, Pei Zhang, Peng Wang, Ru~Peng, Rui Men, Ruize Gao, Runji Lin, Shijie Wang, Shuai Bai, Sinan Tan, Tianhang Zhu, Tianhao Li, Tianyu Liu, Wenbin Ge, Xiaodong Deng, Xiaohuan Zhou, Xingzhang Ren, Xinyu Zhang, Xipin Wei, Xuancheng Ren, Xuejing Liu, Yang Fan, Yang Yao, Yichang Zhang, Yu~Wan, Yunfei Chu, Yuqiong Liu, Zeyu Cui, Zhenru Zhang, Zhifang Guo, and Zhihao Fan. 2024.
\newblock \href {https://doi.org/10.48550/arXiv.2407.10671} {Qwen2 {Technical} {Report}}.
\newblock ArXiv:2407.10671 [cs].

\end{thebibliography}

\section{Language Resource References}
\label{lr:ref}
\bibliographystylelanguageresource{lrec2026-natbib}
\bibliographylanguageresource{languageresource}

\appendix
\clearpage

\section{Human Annotation Instructions}
\label{app:human_annotation_instructions}
\begin{figure*}[ht]
    \centering
     \includegraphics[width=\textwidth, trim=50 110 50 50, clip]{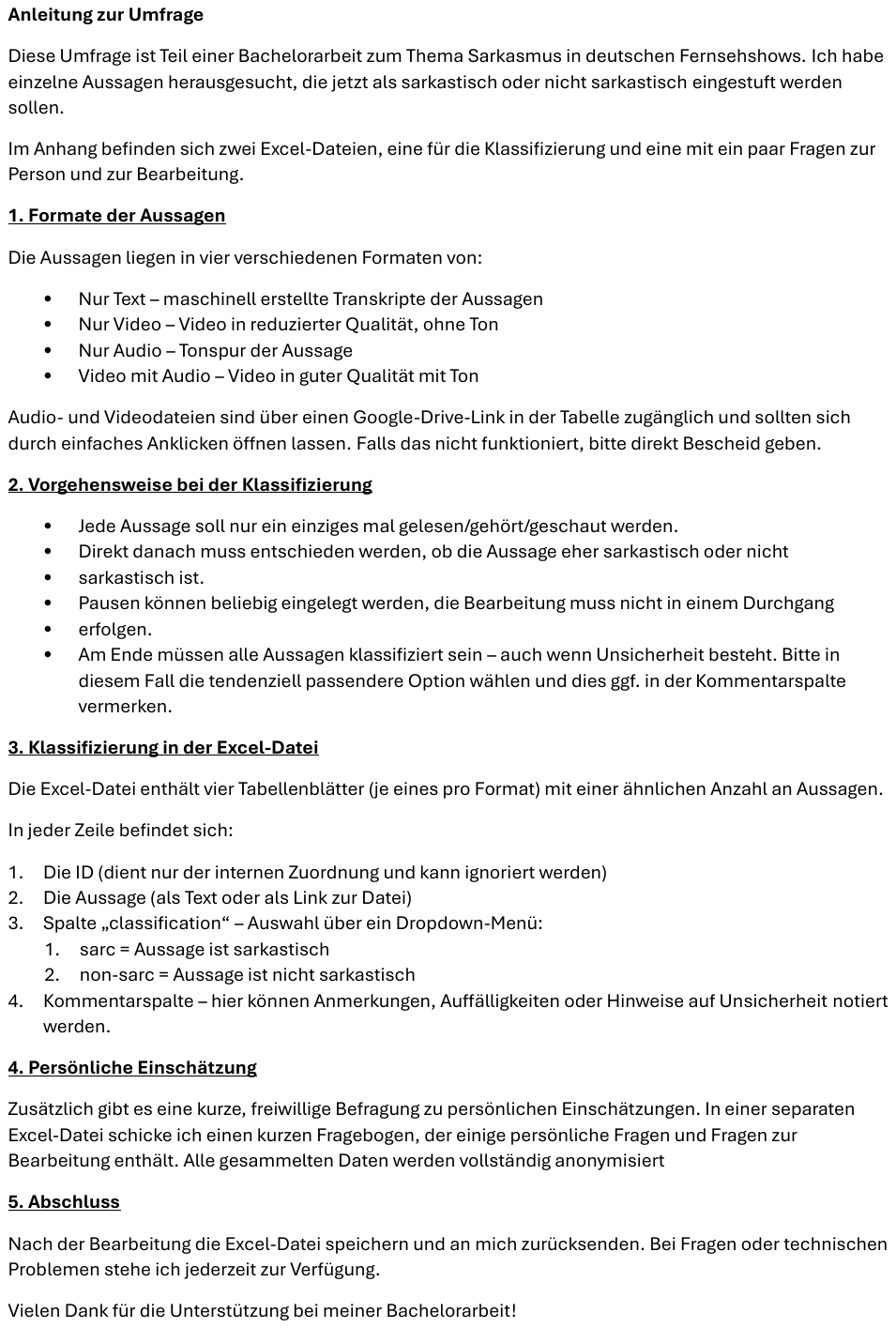}
    \caption{Instructions for human annotators in German.}
    \label{fig:instructions_german}
\end{figure*}

\begin{figure*}[ht]
    \centering
    \includegraphics[width=\textwidth, trim=50 170 50 50, clip]{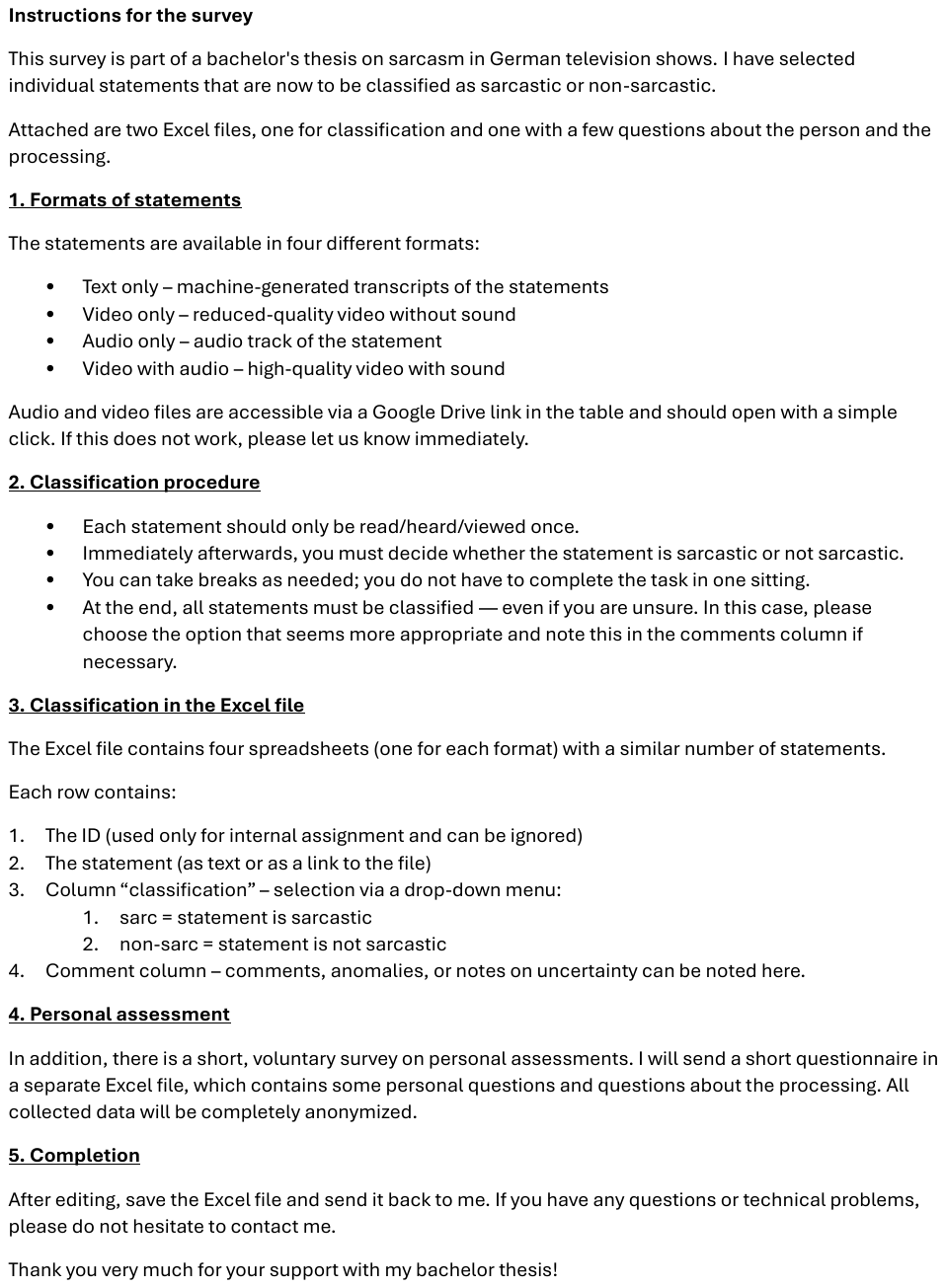}
    \caption{Instructions for human annotators in English, for reference. During the annotation process, the German instructions were used.}
    \label{fig:instructions_english}
\end{figure*}

Detailed instructions for the human annotations can be found in \cref{fig:instructions_german}. 
For reference, we also provide the English translations in \cref{fig:instructions_english}, these have not been used in the annotation process.

\section{Technical Details}\label{app:result_config}

\subsection{General Prompts}
We experiment with different prompts for sarcasm detection. The exact prompts are listed in \cref{fig:prompts1}, \cref{fig:prompts2} and  \cref{fig:prompts3}.

\begin{figure*}[ht]
\textbf{General Prompt:}
\small{\begin{lstlisting}[breaklines=true, breakindent=0pt]
Decide based on the input, if the given utterance is sarcastic or not sarcastic.
Answer ONLY with 'sarc' or 'non-sarc'.
\end{lstlisting}}
\normalsize\textbf{Modality Specific Prompt - text:}
\small{\begin{lstlisting}[breaklines=true, breakindent=0pt]
Decide based on the input, if the given utterance is sarcastic or not sarcastic.
Answer ONLY with 'sarc' or 'non-sarc'.
You will be given ONE short sentence at a time, containing the transcript of a spoken statement.
Examples:
Input: Na das läuft ja mal wieder super!
Output: sarc
Input: Heute morgen war das Wetter eher schlecht.
Output: non-sarc
ANSWER ONLY WITH 'sarc' OR 'non-sarc'!
\end{lstlisting}}
\normalsize\textbf{Modality Specific Prompt - audio:}
\small{\begin{lstlisting}[breaklines=true, breakindent=0pt]
Decide based on the input, if the given utterance is sarcastic or not sarcastic.
Answer ONLY with 'sarc' or 'non-sarc'.
You will receive ONE short audio clip at a time with NO transcript or video available.
Base your decision solely on the audio recording of their speech.
Use ONLY vocal cues such as tone, pitch, pacing, rhythm, stress, intonation, and prosody to decide if the speakers intent is sarcastic.
Examples:
Audio example 1: Speaker uses exaggerated rising intonation and slow pacing in a positive phrase that sounds insincere  
Output: sarc
Audio example 2: Speaker uses normal pitch, steady pacing, and neutral tone in a factual statement  
Output: non-sarc
ANSWER ONLY WITH 'sarc' OR 'non-sarc'!
\end{lstlisting}}
\normalsize\textbf{Modality Specific Prompt - video:}
\small{\begin{lstlisting}[breaklines=true, breakindent=0pt]
Decide based on the input, if the given utterance is sarcastic or not sarcastic.
Answer ONLY with 'sarc' or 'non-sarc'.
You do NOT have access to audio or transcript.
Base your  decision solely on the video of the speaker while speaking.
Focus exclusively on visual sarcasm cues such as:
- Facial expressions (e.g., smirks, raised eyebrows, eye rolls)
- Gestures and hand movements
- Body language and posture
Use these visual signals to decide if the speakers intent is sarcastic.
Examples:
Input Video: Speaker rolls eyes and smirks while speaking  
Output: sarc
Input Video: Speaker maintains neutral expression and relaxed posture  
Output: non-sarc
ANSWER ONLY WITH 'sarc' OR 'non-sarc'!
\end{lstlisting}}
    \caption{Prompts for multimodal sarcasm detection, single-modality.}
    \label{fig:prompts1}
\end{figure*}

\begin{figure*}[ht]
\normalsize\textbf{Modality Specific Prompt - audio-text:}
\small{\begin{lstlisting}[breaklines=true, breakindent=0pt]
Decide based on the input, if the given utterance is sarcastic or not sarcastic.
Answer ONLY with 'sarc' or 'non-sarc'.
You are provided with two inputs:
1. A transcript of the spoken text (analyze phrasing, irony, exaggeration, and contradiction)
2. An audio recording of the speech (analyze prosody, tone, pitch, pacing, and intonation)
Use BOTH inputs together to decide if the speaker's intent is sarcastic.
Examples:
Input Transcript: Na das Wetter ist ja mal wieder super!  
Input Audio: Speaker uses slow pacing and exaggerated rising intonation  
Output: sarc
Input Transcript: Heute morgen war das Wetter eher schlecht. 
Input Audio: Speaker uses neutral tone and steady pacing  
Output: non-sarc
ANSWER ONLY WITH 'sarc' OR 'non-sarc'!
\end{lstlisting}}
\normalsize\textbf{Modality Specific Prompt - video-text:}
\small{\begin{lstlisting}[breaklines=true, breakindent=0pt]
Decide based on the input, if the given utterance is sarcastic or not sarcastic.
Answer ONLY with 'sarc' or 'non-sarc'.
You are provided with two kinds of input:
1. A transcript of the spoken text (what is said)
2. A video of the speaker while speaking (how it is said)
Analyze the transcript for linguistic cues such as irony, contradiction, exaggeration, and phrasing.
Simultaneously analyze the video for visual cues including facial expressions (e.g., smirks, raised eyebrows, eye rolls), gestures, posture, and body language that typically signal sarcasm.
Use BOTH modalities together to make your judgment.
Examples:
Input Transcript: Na das Wetter ist ja mal wieder super! 
Input Video: Speaker rolls eyes and smirks while saying the sentence  
Output: sarc
Input Transcript: Heute morgen war das Wetter eher schlecht.
Input Video: Speaker maintains neutral facial expression and relaxed posture  
Output: non-sarc
ANSWER ONLY WITH 'sarc' OR 'non-sarc'!
\end{lstlisting}}
\normalsize\textbf{Modality Specific Prompt - audio-video:}
\small{\begin{lstlisting}[breaklines=true, breakindent=0pt]
Decide based on the input, if the given utterance is sarcastic or not sarcastic.
Answer ONLY with 'sarc' or 'non-sarc'.
You are provided with two inputs:
1. An audio recording of the speaker (analyze prosody, tone, pitch, pacing, intonation)
2. A video recording of the speaker (analyze facial expressions such as smirks, raised eyebrows, eye rolls, and body language)
Use BOTH audio and video cues together to judge if the speakers intent is sarcastic.
Examples:
Audio: The speaker uses exaggerated rising intonation and slow pacing on a positive phrase  
Video: The speaker smirks and raises eyebrows while speaking  
Output: sarc
Audio: The speaker uses neutral tone and steady pacing  
Video: The speaker maintains relaxed posture and neutral facial expression  
Output: non-sarc
ANSWER ONLY WITH 'sarc' OR 'non-sarc'!
\end{lstlisting}}
    \caption{Prompts for multimodal sarcasm detection, different modality combinations.}
    \label{fig:prompts2}
\end{figure*}

\begin{figure*}[ht]
\normalsize\textbf{Modality Specific Prompt -audio-video-text:}
\small{\begin{lstlisting}[breaklines=true, breakindent=0pt]
Decide based on the input, if the given utterance is sarcastic or not sarcastic.
Answer ONLY with 'sarc' or 'non-sarc'.
You are provided with three inputs:
1. A transcript of the spoken text (analyze linguistic cues such as irony, exaggeration, contradiction)
2. An audio recording of the speech (analyze prosody, tone, pitch, pacing, and intonation)
3. A video of the speaker while speaking (analyze facial expressions like smirks, raised eyebrows, eye rolls, as well as gestures and body language)
Use all three modalities together to accurately judge if the speakers intent is sarcastic.
Examples:
Transcript: Na das Wetter ist ja mal wieder super!
Audio: Speaker uses exaggerated rising intonation and slow pacing  
Video: Speaker smirks and raises eyebrows while speaking  
Output: sarc
Transcript: Heute morgen war das Wetter eher schlecht.
Audio: Speaker uses neutral tone and steady pacing  
Video: Speaker maintains relaxed posture and neutral facial expression  
Output: non-sarc
ANSWER ONLY WITH 'sarc' OR 'non-sarc'!
\end{lstlisting}}

    \caption{Prompts for multimodal sarcasm detection, including all modalities.}
    \label{fig:prompts3}
\end{figure*}

\subsection{Extended context}
\label{apdx:extended_context}

\begin{figure*}[ht]
\normalsize\textbf{Modality Specific Prompt -audio-video-text}
\small{\begin{lstlisting}[breaklines=true, breakindent=0pt]
You are an expert at detecting sarcasm in text data. 
Your task is to classify a TARGET STATEMENT based on the isolated text data of the statement.
To do this, you are provided with the text data of the TARGET STATEMENT including up to 15 seconds of leading CONVERSTAION CONTEXT.
For identification of the statement to classify, the TARGET STATEMENT is again cited in text form.
Use the CONVERSATIONAL CONTEXT only to interpret the target; do not classify the context itself. 
Examples:
Context: We've been stuck in traffic for an hour. Oh great, perfect timing for a road trip.
Target: Oh great, perfect timing for a road trip.
Output: sarc
Context: I finished my report early today.
Target: I finished my report early today.
Output: non-sarc
Context: We forgot the keys. That was an absolutely brilliant idea.
Target: That was an absolutely brilliant idea.
Output: sarc
Context: The sun rises in the east.
Target: The sun rises in the east.
Output: non-sarc
Classify the TARGET STATEMENT using ONLY the provided text data!
Classify ONLY the TARGET STATEMENT as 'sarc' (sarcastic) or 'non-sarc'.
ANSWER ONLY WITH 'sarc' OR 'non-sarc'!
\end{lstlisting}}

    \caption{Prompts for multimodal sarcasm detection, including all modalities.}
    \label{fig:prompts_context}
\end{figure*}

For extended context, we modify the prompt to specify for which utterance the label should be provided, as in \cref{fig:prompts_context}.

\subsection{Inference Parameters}
\cref{tab:configs} lists the generation parameters for different configurations. 
\begin{table}[ht]
    \centering
    \resizebox{\linewidth}{!}{%
    \begin{tabular}{ccccc}
    \toprule
         Config & max new tokens* & sampling & beams & temp  \\
         \midrule
         1 & 3 & true & 2 & 0.7 \\
         2 & 3 & true & 2 & 1.8 \\
         3 & 3 & false & 1 & - \\
    \bottomrule
    \multicolumn{5}{l}{* For Gemini, we set max new tokens to 5.} \\
    \multicolumn{5}{l}{* When enabling thinking, we set max new tokens to 2000.}\\

    \end{tabular}%
    }
    \caption{Different Generation Configurations.}
    \label{tab:configs}
\end{table}

We use the best performing of the three configurations for each model. This mapping is given in \cref{tab:result_configurations}.
\begin{table*}[ht]
    \centering
    \resizebox{\linewidth}{!}{%
    \begin{tabular}{lllll}
    \toprule
    \textbf{Model} & \textbf{Modality} & \textbf{Transformers version} & \textbf{Generatrion} & \textbf{Prompt} \\
    \midrule
    \textmod{} Qwen3-8B & \textmod{} & 4.55.2 & config 2 & modality specific \\
   \textmod{} Qwen2.5-7B-Instruct & \textmod{} & 4.55.2 & config 3 & modality specific \\
    \textmod{} Qwen2-7B-Instruct & \textmod{} & 4.55.2 & config 3 & modality specific \\
    \midrule
    \multirow{2}{*}{\audiomod{} Qwen2-Audio-7B-Instruct} & \audiomod{} & 4.55.2 & config 3 & general \\
    & \audiomod{}\textmod{} & 4.55.2 & config 3 & modality specific \\
    \midrule
    \multirow{2}{*}{\videomod{} Qwen2-VL-7B-Instruct} & \videomod{} & 4.56.0.dev0 & config 3 & general \\
    & \videomod{}\textmod{} & 4.56.0.dev0 & config 1 & general \\
    \multirow{2}{*}{\videomod{}Qwen2.5-VL-7B-Instruct} & \videomod{} & 4.56.0.dev0 & config 3 & general \\
    & \videomod{}\textmod{} & 4.56.0.dev0 & config 3 & general \\
    \midrule
    \multirow{7}{*}{\multimodal{} Phi-4-multimodal-instruct} 
    & \textmod{} & 4.48.2 & config 2 & modality specific \\
    & \audiomod{} & 4.48.2 & config 1 & modality specific \\
    & \videomod{} & 4.48.2 & config 2 & modality specific \\
    & \audiomod{}\textmod{} & 4.48.2 & config 3 & general \\
    & \videomod{}\textmod{} & 4.48.2 & config 2 & modality specific \\
    & \audiomod{}\videomod{} & 4.48.2 & config 3 & general \\
    & \audiomod{}\textmod{}\videomod{} & 4.48.2 & config 1 & general \\
    \midrule
    \multirow{7}{*}{\multimodal{} Qwen2.5-Omni-7B} 
    & \textmod{} & 4.52.3 & config 2 & modality specific \\
    & \audiomod{} & 4.52.3 & config 1 & modality specific \\
    & \videomod{} & 4.52.3 & config 3 & modality specific \\
    & \audiomod{}\textmod{} & 4.52.3 & config 1 & modality specific \\
    & \videomod{}\textmod{} & 4.52.3 & config 2 & modality specific \\
    & \audiomod{}\videomod{} & 4.52.3 & config 1 & modality specific \\
    & \audiomod{}\textmod{}\videomod{} & 4.52.3 & config 2 & modality specific \\
    \midrule
    \multirow{7}{*}{\multimodal{} Gemini-2.5-flash} 
    & \textmod{} & n.s. & config 3 & modality specific \\
    & \audiomod{} & n.s. & config 3 & modality specific \\
    & \videomod{} & n.s. & config 3 & modality specific \\
    & \audiomod{}\textmod{} & n.s. & config 3 & modality specific \\
    & \videomod{}\textmod{} & n.s. & config 3 & modality specific \\
    & \audiomod{}\videomod{} & n.s. & config 3 & modality specific \\
    & \audiomod{}\textmod{}\videomod{} & n.s. & config 3 & modality specific \\
    \bottomrule
    \end{tabular}%
    }
    \caption{The model configurations and prompts that returned the results presented in \cref{tab:results}.
    Since we access Gemini-2.5-flash through the API, there is no transformer version specified.}
    \label{tab:result_configurations}
\end{table*}


\section{Results \textsc{MuSaG-FullAgreement}}
\label{app:results_sd_100}
\begin{table*}[ht]
    \begin{tabular}{llccc}
    \toprule
    \textbf{Modality} & \textbf{Model} & \textbf{Precision} & \textbf{Recall} & \textbf{F1} \\
    \midrule
    N/A & random baseline & 49.34 & 49.31 & 49.19\\
    \midrule
    \multirow{6}{*}{text \textmod{}} 
     & \textmod{} Qwen2-7B-Instruct             & 82.59 & 69.30 & 70.01 \\ 
     & \textmod{} Qwen2.5-7B-Instruct           & 78.2 & 72.95 & 74.03 \\
     & \textmod{} Qwen3-8B                      & \textbf{87.55} & \textbf{88.01} & \textbf{87.76} \\
     & \multimodal{} Phi-4-multimodal-instruct  & 69.13 & 68.72 & 68.9 \\
     & \multimodal{} Qwen2.5-Omni-7B            & 79.3 & 63.37 & 62.31 \\
     & \multimodal{} Gemini-2.5-flash           & 85.13 & 87.17 & 84.37 \\
    
    \midrule 
     \multirow{4}{*}{audio \audiomod{}}     
     & \audiomod{} Qwen2-Audio-7B- Instruct     & 59.31 & 58.97 & 54.79 \\ 
     & \multimodal{} Phi-4-multimodal-instruct  & 55.54 & 54.03 & 45.68 \\
     & \multimodal{} Qwen2.5-Omni-7B            & \textbf{83.81} & 68.12 & \textbf{68.50} \\
     & \multimodal{} Gemini-2.5-flash           & 76.82 & \textbf{73.44} & 66.83 \\ 
     
     \midrule 
     \multirow{5}{*}{video \videomod{}} 
     & \videomod{} Qwen2-VL-7B-Instruct         & 59.96 & 60.46 & 58.50 \\
     & \videomod{} Qwen2.5-VL-7B-Instruct       & \textbf{61.60} & 53.51 & 37.48 \\
     & \multimodal{} Phi-4-multimodal-instruct  & 19.03 & 50.00 & 27.57 \\
     & \multimodal{} Qwen2.5-Omni-7B            & 61.42 & \textbf{60.47} & 55.30 \\
    & \multimodal{} Gemini-2.5-flash            & 59.34 & 59.87 & \textbf{59.10} \\
     \midrule
     \multirow{4}{*}{text-audio \textmod{}\audiomod{}} 
     & \audiomod{} Qwen2-Audio-7B- Instruct     & 62.5 & 62.42 & 58.71 \\ 
     & \multimodal{} Phi-4-multimodal-instruct  & 46.74 & 49.24 & 31.77 \\
     & \multimodal{} Qwen2.5-Omni-7B            & 83.44 & 67.28 & 67.41 \\
     & \multimodal{} Gemini-2.5-flash           & \textbf{91.55} & \underline{\textbf{93.75}} & \underline{\textbf{92.05}} \\
     \midrule
     \multirow{5}{*}{text-video \textmod{}\videomod{}}    
     & \videomod{} Qwen2-VL-7B-Instruct         & 69.70 & 64.61 & 64.88 \\
     & \videomod{} Qwen2.5-VL-7B-Instruct       & 76.77 & 77.39 & 77.02 \\
     & \multimodal{} Phi-4-multimodal-instruct  & 62.59 & 52.31 & 34.04 \\
     & \multimodal{} Qwen2.5-Omni-7B            & 80.31 & 65.06 & 64.62 \\
    & \multimodal{} Gemini-2.5-flash            & \textbf{88.11} & \textbf{89.19} & \textbf{88.54} \\
     \midrule 
     \multirow{3}{*}{audio-video \audiomod{}\videomod{}} 
     & \multimodal{} Phi-4-multimodal-instruct  & 58.41 & 51.95 & 34.47 \\
     & \multimodal{} Qwen2.5-Omni-7B            & \textbf{82.15} & 68.45 & 68.97 \\
     & \multimodal{} Gemini-2.5-flash           & 79.44 & \textbf{79.8} & \textbf{79.61} \\
     \midrule
     \multirow{3}{*}{text-audio-video \textmod{} \audiomod{}\videomod{}} 
     & \multimodal{} Phi-4-multimodal-instruct  & 61.41 & 59.56 & 52.98 \\
     & \multimodal{} Qwen2.5-Omni-7B            & 84.68 & 73.53 & 74.95 \\
     & \multimodal{} Gemini-2.5-flash           & \underline{\textbf{91.79}} & \textbf{91.79} & \textbf{91.79} \\
     \bottomrule
    \end{tabular}
    \caption{Results on \textsc{MuSaG-FullAgree}. For each modality, we report the same modality-specific models (\textmod{}, \audiomod{}, \videomod{}) and multimodal models (\multimodal{}).}
    \label{tab:results_sd_100}
\end{table*}
We report results for all models on  \textsc{MuSaG-FullAgreement} in \cref{tab:results_sd_100}. In \cref{tab:sd100_and_human} in the main paper, we also report human evaluation on different modalities.

\section{Examples Predictions for Extended Context}\label{app:examples}

\cref{tab:context_effects} illustrates how extended context influences model predictions. We show three cases: examples where additional context \textit{improved} classification (all models corrected their prediction), where it \textit{degraded} classification (previously correct models were misled by the surrounding utterances), and where context had \textit{no impact} (models predicted correctly regardless). In the \textit{With context} rows, the target utterance is highlighted in \textbf{bold}.


\begin{table*}[tp]
\centering
\resizebox{\textwidth}{!}{%
\renewcommand{\arraystretch}{1.4}
\begin{tabular}{l l p{7cm} c p{3.5cm}}
\toprule
\textbf{Effect} & \textbf{Setting} & \textbf{Transcript} & \textbf{True Label} & \textbf{Model Preditions} \\
\midrule

Improved
& Without context &
\textit{„Genau so und nur so gewinnt man die Herzen der Menschen."}
& \lblsarc &
\begin{tabular}[t]{@{}l@{}}
Text: \lblnonsarc \\ Audio: \lblnonsarc \\ Video: \lblnonsarc \\ AV: \lblnonsarc
\end{tabular} \\[4pt]

& With context &
\textit{„[\ldots] Wahlprogramm nicht gelesen haben und dass sie das, was sie gerade sagen, auswendig gelernt haben. Genau das Gegenteil. Ja, offensichtlich. Ja, darf ich mal? Klar darfst du. \textbf{Genau so und nur so gewinnt man die Herzen der Menschen.}"}
& \lblsarc &
\begin{tabular}[t]{@{}l@{}}
Text: \lblsarc \\ Audio: \lblsarc \\ Video: \lblsarc \\ AV: \lblsarc
\end{tabular} \\

\midrule

Degraded
& Without context &
\textit{„Bevor das Gesetz endgültig beschlossen wird, dürfen jetzt erstmal alle Fraktionen im Bundestag draufschauen und Änderungsvorschläge einbringen."}
& \lblnonsarc &
\begin{tabular}[t]{@{}l@{}}
Text: \lblnonsarc \\ Audio: \lblnonsarc \\ Video: \lblnonsarc \\ AV: \lblnonsarc
\end{tabular} \\[4pt]

& With context &
\textit{„Damit sich das ändert, will die Bundesregierung die Bonitätseinschätzung von Auskunftteilen nun stärker reglementieren. Das klingt ja erstmal so, als wäre jetzt alles in trockenen Tüchern, Schufa ausgedribbelt, aber weit gefehlt. Denn dieser Entwurf ist ja erst der Anfang. \textbf{Bevor das Gesetz endgültig beschlossen wird, dürfen jetzt erstmal alle Fraktionen im Bundestag draufschauen und Änderungsvorschläge einbringen.}"}
& \lblnonsarc &
\begin{tabular}[t]{@{}l@{}}
Text: \lblsarc \\ Audio: \lblnonsarc \\ Video: \lblsarc \\ AV: \lblsarc
\end{tabular} \\

\midrule

No impact
& Without context &
\textit{„Haben Sie das auch gehört? Hat Herr Klingbeil gerade gesagt, deutsche Stärken sind moderne Bildung, bezahlbares Wohnen und digitale Infrastruktur? Er war aber schon in Deutschland, oder? Ja."}
& \lblsarc &
\begin{tabular}[t]{@{}l@{}}
Text: \lblsarc \\ Audio: \lblsarc \\ Video: \lblsarc \\ AV: \lblsarc
\end{tabular} \\[4pt]

& With context &
\textit{„[\ldots] ob Lars Klingbeil das hier wirklich ernst meint. Wir setzen auf die Stärke unseres Landes, auf gute Arbeit, auf moderne Bildung, auf bezahlbares Wohnen, auf digitale Infrastruktur, auf klimafreundliche Energie. \textbf{Haben Sie das auch gehört? Hat Herr Klingbeil gerade gesagt, deutsche Stärken sind moderne Bildung, bezahlbares Wohnen und digitale Infrastruktur? Er weiß aber schon in Deutschland, oder? Ja.}"}
& \lblsarc &
\begin{tabular}[t]{@{}l@{}}
Text: \lblsarc \\ Audio: \lblsarc \\ Video: \lblsarc \\ AV: \lblsarc
\end{tabular} \\

\bottomrule
\end{tabular}
}%
\caption{Examples illustrating the effect of extended context on classification. In the \textit{With context} rows, the target sentence is shown in \textbf{bold}. Model predictions are shown per modality: Text\,=\,Qwen3-8B, Audio\,=\,Gemini-2.5-flash (audio), Video\,=\,Gemini-2.5-flash (video), AV\,=\,Gemini-2.5-flash (audio-video).}
\label{tab:context_effects}
\end{table*}
\end{document}